%% file: arxiv_main.tex
\documentclass[10pt,twocolumn,letterpaper]{article}
\usepackage[numbers,sort&compress]{natbib}

\usepackage{iccv}
\usepackage{times}
\usepackage{epsfig}
\usepackage{graphicx}
\usepackage{amsmath}
\usepackage{amssymb}
\usepackage{soul}
\usepackage{xcolor}
\usepackage{booktabs}
\usepackage{pifont}
\usepackage{amsfonts}
\usepackage{bbm}
\usepackage{soul}
\usepackage{float}
\usepackage{multirow}
\usepackage{array,makecell}
\usepackage{enumitem}

\usepackage{hyperref}

\usepackage{xr-hyper}
\hypersetup{
    pagebackref=true,
    breaklinks=true,
    letterpaper=true,
    bookmarks=false,
    colorlinks=true, %
    linktoc=all,     %
}
\usepackage[numbers,sort&compress]{natbib}

\iccvfinalcopy %

\ificcvfinal\fi

\begin{document}

\title{\ours : Audio Conditioned Diffusion Models for Lip-Synchronization}

\author{Soumik Mukhopadhyay$^1$\\
{\tt\small soumik@umd.edu}
\and
Saksham Suri$^1$\\
{\tt\small sakshams@cs.umd.edu}
\and
Ravi Teja Gadde$^2$\\
{\tt\small rtg267@nyu.edu}
\and
Abhinav Shrivastava$^1$\\
{\tt\small abhinav@cs.umd.edu}
\and
$^1$University of Maryland, College Park\\
\and 
$^2$New York University
}

\ificcvfinal\thispagestyle{empty}\fi

\ificcvfinal\pagestyle{empty}\fi

\newcommand{\methodname}{Diff2Lip}
\newcommand{\ours}{\methodname}

\maketitle
\input{0_abstract}

\input{1_intro}

\input{2_related_work}

\input{3_method}

\input{4_experiments}

\input{5_discussion}

{\small
\bibliographystyle{ieee_fullname}
\bibliography{egbib}
}

\clearpage
\appendix

\input{6_supp}

\end{document}

%% file: 0_abstract.tex
\begin{abstract}
The task of lip synchronization~(lip-sync) seeks to match the lips of human faces with different audio.   
It has various applications in the film industry as well as for creating virtual avatars and for video conferencing. 
This is a challenging problem as one needs to simultaneously introduce detailed, realistic lip movements while preserving the identity, pose, emotions, and image quality. Many of the previous methods trying to solve this problem suffer from image quality degradation due to a lack of complete contextual information.
In this paper, we present \ours, an audio-conditioned diffusion-based model which is able to do lip synchronization in-the-wild while preserving these qualities.
We train our model on Voxceleb2, a video dataset containing in-the-wild talking face videos.   
Extensive studies show that our method outperforms popular methods like Wav2Lip and PC-AVS in Fréchet inception distance~(FID) metric and Mean Opinion Scores~(MOS) of the users. We show results on both reconstruction (same audio-video inputs) as well as cross (different audio-video inputs) settings on Voxceleb2 and LRW datasets. Video results and code can be accessed from our \href{https://soumik-kanad.github.io/diff2lip}{project page}.

\end{abstract}

\vspace{-0.2in}

%% file: 1_intro.tex
\section{Introduction}

\begin{figure}[!t]
    \centering
    \includegraphics[width=\linewidth]{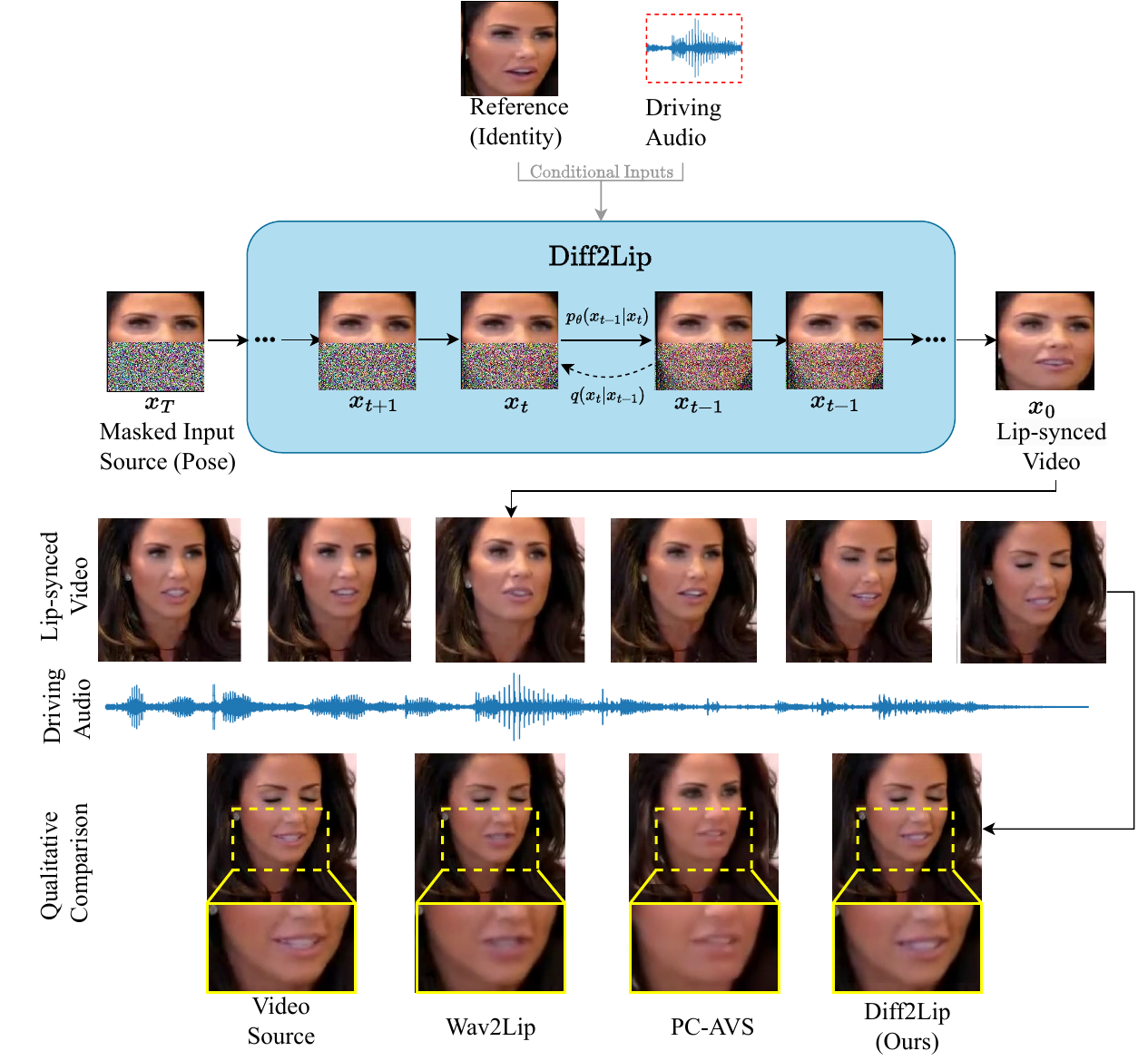}
    \caption{\textbf{Top:} Our \ours~~approach uses an audio-conditioned diffusion model to generate lip-synchronized videos. (Here $q$ denotes the forward diffusion process and $p_\theta$ is the learned reverse diffusion process.) \textbf{Bottom:} On zooming in to the mouth region it can be seen that our method generates high-quality video frames without suffering from identity loss.}
    \label{fig:teaser}
    \vspace{-0.2in}
\end{figure}

Oscar-winning director Bong Joon Ho famously pointed out that subtitles act as a barrier between a foreign audience not adept in the language and their ability to fully enjoy amazing movies~\cite{subtitle-quote}, as the viewer needs to focus on both watching and reading. A rarely explored alternative, multiple-language version movie (MLV), where the same film is shot in multiple languages in parallel, is naturally much more expensive~\cite{mlv}. While dubbing is a popular compromise solution, it can feel unnatural due to the lack of synchronization between speech and actors' video. 
As a cheaper alternative, lip-synchronization (lip-sync) aims to generate the mouth region of the human face such that the lips correspond to a different speech audio. Its applications beyond movies include education, virtual avatars, and video conferencing.  
Ideally, lip-sync should support any identity and audio from unseen sources (in-the-wild setting). 
This setting brings up the challenges of preserving the actors’ identity, pose, emotions, and visual quality while maintaining a realistic lip-sync.

One of the earliest lip-sync methods, Video Rewrite~\cite{video-rewrite}, had a purpose-built solution by mapping phonemes to mouth shapes and then blending them onto the target video. Modern techniques have more general solutions but suffer certain limitations. 
For instance, PC-AVS~\cite{pcavs} and GC-AVT~\cite{gcavt}, disentangle pose and expression respectively but fail to preserve identity (Fig.~\ref{fig:teaser} bottom), have worse visual quality, and have border inconsistencies (while putting the generated heads back to the scene). On the other hand, works that target a specific identity, such as SynthesizingObama~\cite{synthesizing-obama}, require video/identity-specific training. Other methods which rely on extracting intermediate representations, e.g., landmarks in MakeItTalk~\cite{makeittalk}, have to deal with estimation errors in these representations. Finally, approaches that can generalize on in-the-wild lip-sync settings pose it as an inpainting task, where the mouth region is masked and then generated according to the audio. Examples include Wav2Lip~\cite{wav2lip}, which achieves good lip-sync but at the cost of poor visual quality (see Fig.~\ref{fig:teaser} bottom), and AV-CAT~\cite{avcat}, which has a multistage pipeline but does not capture finer details. In this paper, we introduce \ours, an inpainting style approach that solves the lip-sync task using diffusion models, which addresses most of these shortcomings and achieves visually superior lip-sync results.

We propose an audio-conditioned diffusion model to solve the task of lip-sync (Fig.~\ref{fig:teaser} top). 
Diffusion models~\cite{ddpm} are likelihood-based models that can generate astonishing results in high variation datasets~(e.g., ImageNet~\cite{imagenet}), that GANs~\cite{gan} cannot match. To generalize in-the-wild, we pose the problem as a conditional diffusion model based inpainting task~\cite{palette}. \ours  ~takes three inputs: a masked input frame, a reference frame, and an audio frame, and outputs the lip-synced mouth region. \ours  ~leverages (1) the masked input frame to get the pose context; (2) the reference frame to get the identity and mouth region textures; (3) the audio frame to drive the lip shape.
Using an audio+image conditioned diffusion model, \ours  ~maintains a fine balance between all these contextual input information, avoiding lip-sync problems (e.g. identity loss, reference copying, inaccurate lip shape). \ours  ~optimizes three losses: a reconstruction loss to guide synthesis; a sync-expert loss~\cite{wav2lip} to enforce synchronization; and a sequential adversarial loss to enforce inter-frame continuity.
\ours  ~generates high image quality without identity loss or generalizability issues as shown in Fig.~\ref{fig:teaser} bottom.

We evaluate our work on commonly used benchmarks of Voxceleb2~\cite{voxceleb2} and LRW~\cite{lrw} datasets for the tasks of reconstruction and cross generation (see section~\ref{sec:exp}). We compare against popular methods used for lip-sync like Wav2Lip~\cite{wav2lip} and PC-AVS~\cite{pcavs}. 
Extensive evaluations show that \ours ~outperform existing methods in terms of image fidelity while having comparable synchronization.

The following are the contributions of this work:
\begin{itemize}[noitemsep,topsep=0pt]
    \item  
    We propose a novel diffusion model based approach for audio-conditioned image generation.
    \item Using frame-wise and sequential losses we are able to successfully generate high quality lip-sync.
    \item We show that the use of a sequential adversarial loss makes frame-wise video generation more stable for diffusion models across frames.
    \item Extensive evaluations validate that our generations outperform existing methods in FID metric and MOS of the users showing the effectiveness of \ours. 
\end{itemize}

%% file: 2_related_work.tex
\section{Related Works}
In this section, we first talk about existing methods in lip-sync and then discuss conditional diffusion models. 
\subsection{Lip synchronization}
Lip-sync methods can be roughly classified into the following four categories.  Please note that there may be overlaps between these categories.

\noindent{\textbf{Embedding-based head reconstruction.}}
\label{sec:head-based}
This class of methods tends to synthesize the entire head by the fusion of speech and identity features. This is usually done using an encoder-decoder style architecture. Song et al.~\cite{song2018talking} and Vougioukas et al.~\cite{vougioukas2018end} use RNNs while Speech2Vid~\cite{speech2vid} uses CNNs. LipGAN~\cite{kr2019towards} uses an audio-visual discriminator to improve synchronization. PC-AVS~\cite{pcavs} performs disentanglement of identity, speech, and pose from each other to have complete pose control. GC-AVT~\cite{gcavt} additionally disentangles emotion. Recent contemporary works like DiffTalk~\cite{shen2023difftalk} use latent diffusion models for achieving high visual quality at the cost of lip-sync, which is even worse in cross generation. This method additionally requires landmarks for proper face positioning, uses auto-regressive inference strategies that cannot be parallelized, and employs an external frame interpolation method as it suffers from jitter. In general, as these methods generate full faces, they suffer from border inconsistency issues while putting the generated head back onto the frame.

\noindent{\textbf{Intermediate representation-based methods.}}
\label{sec:2d3d}
These methods learn to manipulate sparse intermediate representations like face landmarks or meshes. Chen et al.~\cite{chen2019hierarchical} and Das et al.~\cite{das2020speech} generate faces conditioned on the landmarks estimated using the audio. MakeItTalk~\cite{makeittalk} proposes to predict speaker landmark displacement based on the audio. 
Methods like Song et al.~\cite{song2022everybody}, Yu et al.~\cite{yu2020multimodal}, and Xie et al.~\cite{xie2021towards} use 3DMM (facial mesh) to generate face videos. Neural Voice Puppetry~\cite{thies2020neural}, uses audio to predict the coefficients of an expression basis of a 3D model. Zhang et al.~\cite{zhang2021flow} propose to first predict 3DMM-based animation parameters which are then converted into a dense flow for facial animation. Although these methods leverage intermediate structures for lip-sync, getting such representations manually is expensive while automatic predictions are error-prone. Further, these also tend to lose finer details given the sparse representation.

\noindent{\textbf{Personalized methods.}}
\label{sec:person-specific}
In this type of methods, the models are trained to be identity specific or even video-specific~\cite{rhythmic-head}. For example, SynthesizingObama~\cite{synthesizing-obama} only focuses on Obama's lip sync using an audio-to-landmark network.  MEAD~\cite{mead} leverages edges while Lu et al.\cite{lu2021live} uses facial landmarks to create edge-like conditional-feature maps to generate talking faces. Methods like  Song et al.~\cite{song2022everybody}, Zhang et al.\cite{zhang2021flow}, and Neural Voice Puppetry~\cite{thies2020neural}, discussed earlier, which do audio-driven expression manipulation (3DMM) also fall in this category. Some methods also deal with explicit 3D mesh vertex deformfation like LipSync3D~\cite{lipsync3d}. NeRF~\cite{nerf} based models like AD-NeRF~\cite{adnerf} and SSPNeRF~\cite{SSPNeRF} are also person-specific. This class of methods demonstrates high video quality at times, but that comes at the cost of retraining the model on the specific person and environment every time.

\noindent{\textbf{Inpainting-based methods.}}
\label{sec:inpainting-based}
In these methods instead of generating the whole face only the bottom part of the face, which gets affected by speech is modified. These models don't suffer from image boundary inconsistencies when pasting back the mouth portion to the entire frame. Initial works like~\cite{chen2018lip} only focus on lip region features and used an audio-speech fusion module to merge them. Wav2Lip~\cite{wav2lip} is one of the most popular methods in this area and shows the importance of lip-sync expert network for better lip-sync. AV-CAT~\cite{avcat} uses a transformer backbone and a refinement model for inpainting the lower face.  SyncTalkFace~\cite{synctalkface} further introduces an audio lip memory that is used for inference time generation. Our method falls also in this category of lip-sync. It doesn't struggle with error propagation issues being end-to-end trainable, and does not require any explicit 2D/3D information while being identity-agnostic. It further improves the image quality compared to previous methods. Recent contemporary works like Gupta et al.~\cite{gupta2023towards} also focus on improving quality by using VQGAN and a face restoration network but consequently make the lips have a pinkish tint and slightly different from the input. Their lip-sync expert network requires five times more context compared to Wav2Lip.

\subsection{Conditional diffusion models}

Initial work in this area involved class conditioning for image generation. For example, Ho \& Salimans~\cite{ho2021classifierfree} used class labels to train a diffusion model by interpolating between conditional and unconditional outputs. While Guided-diffusion~\cite{guided-diffusion} used a classification network for class conditioning to get a better image generation. Methods like GLIDE~\cite{glide}, DALLE-2~\cite{dalle2}, Stable-Diffusion\cite{latent-diffusion}, and IMAGEN~\cite{imagen} leverage language models to generate photorealistic as well as many other styles of images just using text prompt inputs. There has also been some research into text-to-video generation like Video Diffusion Models~\cite{video-diffusion-models} and more photorealistic models like Gen-1~\cite{gen1}. Recently, seeing the popularity of diffusion models, people have also proposed models like Noise2Music~\cite{noise2music} that can generate music using just text prompts. There has also been some work on image-conditioned diffusion models. Palette~\cite{palette} is a generalized image-to-image generation framework, which can solve tasks like coloration, inpainting, outpainting, jpeg-restoration, etc. We also pose the problem as an inpainting style diffusion task but with additional audio and reference identity-conditioned inputs.

%% file: 3_method.tex
\begin{figure*}[!t]
    \centering
    \includegraphics[width=0.9\linewidth]{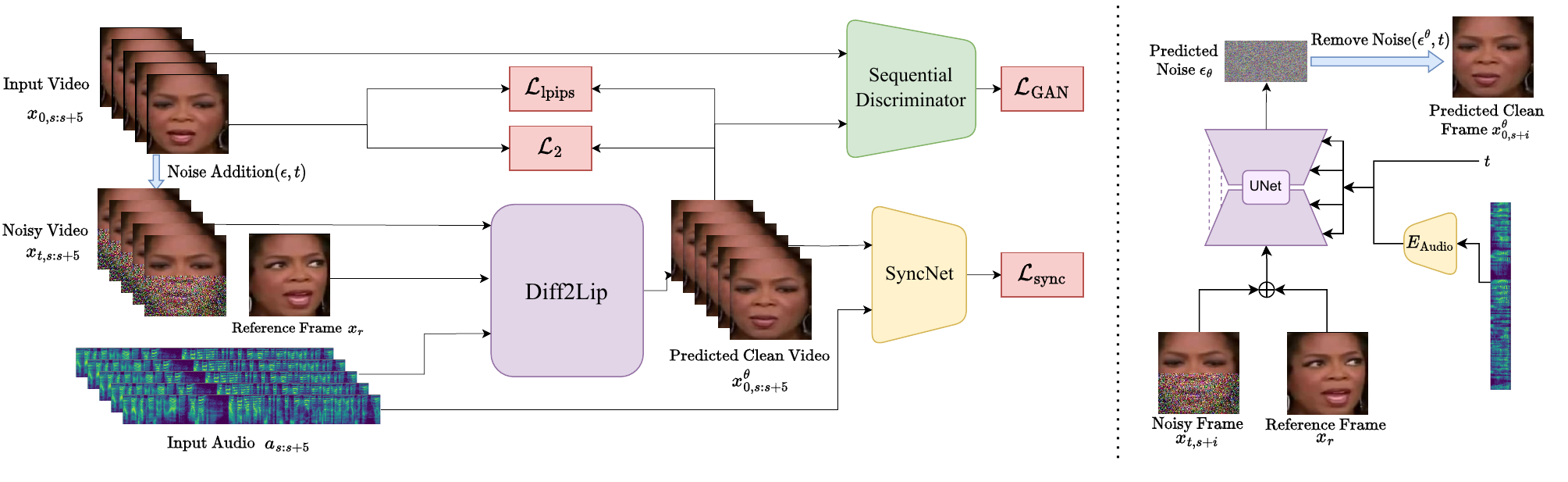}
    \vspace{-0.1in}
    \caption{\textbf{Overview:} \ours ~ solves lip-sync using an audio-conditioned diffusion model, which learns to inpaint the lower half of the face. During training (left), given an input video sequence $x_{0,s:s+5}$, we first add noise to the lower half using the forward process (Eq.~\ref{eq:make_noisy}) to get the noisy video sequence $x_{t,s:s+5}$, where diffusion step $t$ is sampled uniformly. Then a noisy video frame $x_{t,s+i}$ for $i\in[0,5)$, a different random reference frame $x_{r}$, and the audio frame $a_{s+i}$ is input  to our model. The audio encoder $E_\text{Audio}$ encodes the audio frame $a_{s+i}$. Our model (right), predicts the added noise $\epsilon_\theta$ given these inputs, which is used to get the predicted clean frame $x^\theta_{0,s+i}$ (using Eq.~\ref{eq:make_noisy}). Then frame-wise reconstruction losses like $\mathcal{L}_2$ and $\mathcal{L}_\text{lpips}$ are applied to the predicted clean sequence $x^\theta_{0,s:s+5}$ for enforcing good image quality while sequential losses like sequential adversarial loss $\mathcal{L}_\text{GAN}$ and SyncNet expert loss $\mathcal{L}_\text{sync}$ ensure lip-sync.  }
    \label{fig:approach}
    \vspace{-0.1in}
\end{figure*}

\section{Methods}

In this section, we discuss our proposed approach - \ours. We propose a novel audio and image conditioned diffusion model which is able to synthesize high quality lip-synced mouths corresponding to the audio input.  We discuss diffusion models in Section~\ref{sec:dm}. Then we introduce our approach in Section~\ref{sec:approach}. Finally, in Section~\ref{sec:losses}, we talk about the losses required to train our model.

\subsection{Diffusion Models}\label{sec:dm}
Diffusion models~\cite{ddpm} are likelihood-based models which try to sample points from a given distribution by gradually denoizing random gaussian noise in $T$ steps. In the forward diffusion process, increasing amounts of noise is added to a sample point $x_0$ iteratively as $x_0\rightarrow x_1 \rightarrow \dots \rightarrow x_{t-1}\rightarrow x_t \rightarrow \dots \rightarrow x_{T-1}\rightarrow x_T$, to get a completely noisy image $x_T$. Formally, the forward diffusion process is a Markovian noising process defined by a list of noise scales ${\{\Bar{\alpha}_t\}}_{t=1}^T$ as:

\begin{equation}
\label{eq:forward}
q(x_t|x_0) := \mathcal{N}(x_t|\sqrt{\Bar{\alpha}_t}x_0, (1-\Bar{\alpha}_t)\textbf{I})
\end{equation}
which can be rewritten as:

\begin{equation}
\label{eq:make_noisy}
x_t = \sqrt{\Bar{\alpha}_t}x_0 + \sqrt{1-\Bar{\alpha}_t}\epsilon, \space \epsilon \in \mathcal{N}(0, \textbf{I})
\end{equation}
where $\epsilon$ is the noise, $\mathcal{N}$ denotes normal distribution, $x_0$ is the original image, and $x_t$ is noised image after $t$ steps of the diffusion process.  
The reverse diffusion process aims to learn the posterior distribution $q(x_{t-1}|x_0,x_t)$, using which one can estimate $x_{t-1}$ given $x_t$. This is typically done using a neural network, which can be parameterized in multiple ways. Similar to~\cite{ddpm, improved-diffusion, guided-diffusion}, we choose to parameterize the neural network to predict the noise, ie. $\epsilon_\theta(x_t, t)$, where $\theta$ represents the parameters of the neural network. It takes a noisy sample $x_t$ and timestep $t$ to predict the added noise $\epsilon$ in Eq.~\ref{eq:make_noisy}. The model is learned using the simplified objective used in~\cite{ddpm} which reweights the variational lower bound on the maximum likelihood objective: 

\begin{equation}
\label{eq:L_eps}
\mathcal{L}_{\text{simple}}=\mathbb{E}_{x_0,t,\epsilon}[\|\epsilon_\theta(x_t, t) - \epsilon \|_2^2]
\end{equation}

 The posterior distribution $q(x_{t-1}|x_0,x_t)$ is also tractable using the Bayesian rule and turns out to be another normal distribution. When using DDIM~\cite{ddim} for sampling, we can deterministically sample the posterior by disregarding the variance. Since we can write $x_0$ in terms of $x_t$ and $\epsilon$ using Eq.~\ref{eq:make_noisy}, therefore we can recover $x_{t-1}$ deterministically given $x_t$ and $\epsilon$ using:
 \begin{equation}
\label{eq:remove_noise}
x_{t-1}=\sqrt{\frac{\Bar{\alpha}_{t-1}}{\Bar{\alpha}_t}}x_t + \left( \sqrt{1-\Bar{\alpha}_{t-1}} -  \sqrt{\frac{\Bar{\alpha}_{t-1}(1-\Bar{\alpha}_t)}{\Bar{\alpha}_t}}\right)\cdot \epsilon
\end{equation}
This equation represents the mean of the learned posterior $p_\theta(x_{t-1}|x_{x_t})$ distribution in the DDIM~\cite{ddim} formulation.

\begin{figure}[!t]
    \centering
    \includegraphics[width=\linewidth]{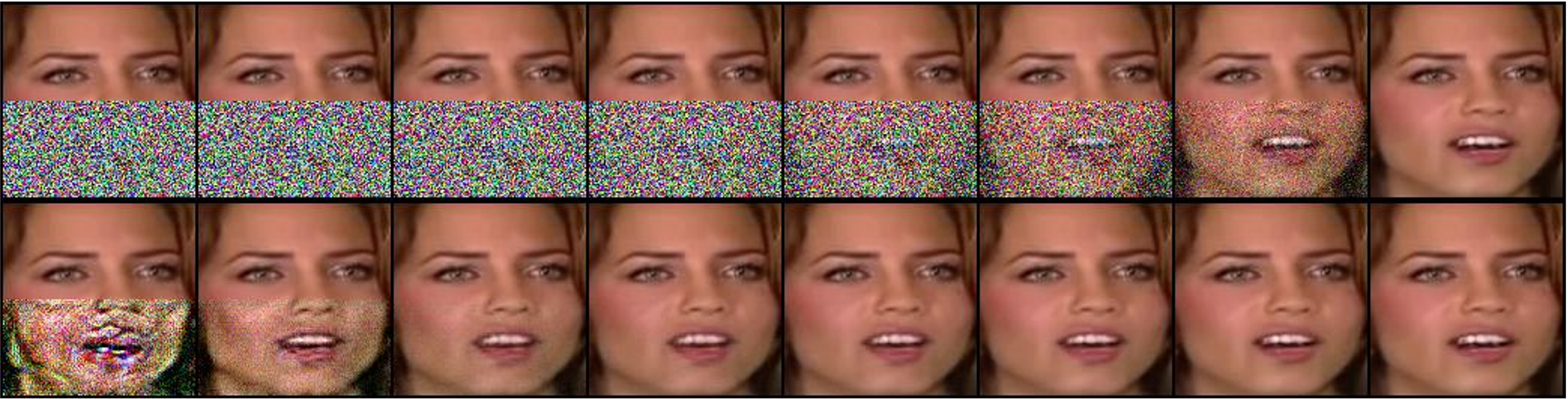}
    \caption{Intermediate $x_t$ (\textbf{top}) and $x^\theta_{0}$ (\textbf{bottom}) as $t$ goes from $T$ to $0$ (left to right), sampled at uniform intervals.}
    \label{fig:intermediate}
    \vspace{-0.15in}
\end{figure}

For sampling during inference time, $x_T$ is sampled from the standard normal distribution. The neural network can then recover the noise $\epsilon_\theta$ that needs to be removed. This in turn can be fed into Eq.~\ref{eq:remove_noise} to get back $x_{T-1}$. Iterating over this one can get the clean image as $x_T\rightarrow x_{T-1} \rightarrow \dots \rightarrow  x_t\rightarrow x_{t-1} \rightarrow \dots \rightarrow x_1\rightarrow x_0$ as seen in Fig.~\ref{fig:intermediate} top.

\textbf{Notation.} In this paper, we work with diffusion processes and videos. We use $t$ for the diffusion process step number while $s$ for the video frame number.  For the diffusion process, we keep the notation here the same as~\cite{improved-diffusion}.

\subsection{Proposed Approach}\label{sec:approach}

We pose the problem of lip-sync as a lower mouth inpainting task,  where given an input face with the lower half masked, an audio frame input, and a reference frame input, the model needs to generate the masked region of the face. Formally, given a video $V=\{v_1,\dots v_S\}$ with $S$ frames, were $v_s$ is the $s^\text{th}$ frame, and audio $A=\{a_1,\dots, a_S\}$, where $a_s$ is the $s^\text{th}$ audio frame, our model processes one video frame $x_{0,s}=v_s$ at a time. Let $x_{s,T}=v_s\cdot (1-M) + \eta \cdot M$ be a noise-masked video frame, where $\eta \in \mathcal{N}(0, \textbf{I})$ and $M$ is a binary mask for the lower half of the face. (Here the subscript $T$ denotes a completely noised frame that we want to denoise). We want our trained model to be able to recover $v_s$ using the reverse diffusion process, given inputs masked video frame $x_{s,T}$, the audio frame $a_s$, and a random reference frame $x_r=v_{\text{random}(1,S)\neq s}$. This setup is quite similar to Wav2Lip\cite{wav2lip}. The random reference frame $x_r$ is chosen from the same video and provides cues about the source's identity and pose. We make sure that it is not the same as the input frame; otherwise there could be information leakage while training. The audio input $a_s$ provides information about the lip structure.  

As shown Fig.~\ref{fig:approach} we formulate the generation as an inpainting task similar to~\cite{palette}, i.e.,  we learn a conditional model $\epsilon_\theta(x_{s,t}, a_s, x_r,t)$. At training time, we first take a clean sample frame $x_{s,0}(=v_s)$ and a uniformly sampled $t$, and add noise to $x_{s,0}$ using Eq.~\ref{eq:make_noisy} to get $x_{s,t}$. The model is trained to predict the noise $\epsilon  \in \mathcal{N}(0, \textbf{I})$ added to it using $\mathcal{L}_{\text{simple}}$ (Eq.~\ref{eq:L_eps}).

We feed the reference frame by concatenating it with the input frame while the audio is fed using group normalization (similar to time and class conditioning in~\cite{improved-diffusion}). Our network has a UNet~\cite{unet} backbone which consists of residual blocks and attention blocks similar to~\cite{guided-diffusion}.  We want the UNet to extract contextual information from the unmasked portion of the input frame, and the reference frame. To enforce this we provide these directly as input to the network. For the audio which is used as a conditioning, we first encode it using a trainable encoder $E_\text{Audio}$, which generates embeddings that are injected as conditioning. $E_\text{Audio}$ is also built using the same blocks as the UNet.

\subsubsection{Additional Losses}\label{sec:losses}

When just training using $\mathcal{L}_{\text{simple}}$ (applied to the masked region), we observe that the mouth region generation had good image quality but no lip-sync.  
Hence we add additional losses to make our model work.

Our model predicts in noise space and hence many image-space losses cannot be directly applied to it. There are three ways to approach this issue - first, our model could be parameterized to directly predict the denoized image $x_0$ instead of predicting $\epsilon$. Second, we can use the sampling process described in Section~\ref{sec:dm} to recover back the clean image $x_0$. Third, substituting $x_t$ and $\epsilon_\theta$ into Eq.~\ref{eq:make_noisy} one could directly recover $x^\theta_0(x_t, t)$, an estimate of $x_0$, without having to do iterative sampling. We observed that directly predicting denoized image leads to worse image quality while using iterative sampling is overly time-consuming and hence we stick with predicting $\epsilon$.

This approach leads to a noisy $x^\theta_0(x_t, t)$ when the step $t$ is large as seen in Fig.~\ref{fig:intermediate} bottom but there have been previous works~\cite{piti} which have applied image losses to $x^\theta_0(x_t, t)$. We enforce an $\mathcal{L}_2$ loss on this $x^\theta_0(x_t, t)$ to make sure that it is clean:
\begin{equation}
\label{eq:L_2}
\mathcal{L}_{2}=
\mathbb{E}_{x_{0,s},t,\epsilon}[\|x^\theta_{0,s} - x_{0,s} \|_2^2]
\end{equation}
Next, to impose audio synchronization, we use SyncNet discriminator as used by Wav2Lip~\cite{wav2lip}. We first separately train the SyncNet in a contrastive manner which is kept fixed during training our generation model. Similar to Wav2Lip~\cite{wav2lip} we work with a sequence of 5 frames as input to the SyncNet. By using 5 predicted video frames $x^\theta_{0,s:s+5}$ and the corresponding audio sequence $a_{s:s+5}$, SyncNet loss can be written as:
\begin{equation}
\label{eq:L_sync}
\mathcal{L}_{\text{sync}}=\mathbb{E}_{x_{0,s},t,\epsilon}[\text{SyncNet}(x^\theta_{0,s:s+5}, a_{s:s+5} )]
\end{equation}

As shown in our ablation, directly adding SyncNet loss, deteriorates the image quality. To mitigate this we add perceptual similarity loss~\cite{lpips} on the generated frames:
\begin{equation}
\label{eq:L_lpips}
\mathcal{L}_{\text{lpips}}=\mathbb{E}_{x_{0,s},t,\epsilon}\mathbb{E}_{l}[ \|\phi_l(x^\theta_{0,s})-\phi_l(x_{0,s} )\|_2^2]
\end{equation}
where $\phi_l(\cdot)$ represents the features coming from the $l^\text{th}$ layer of a pretrained-VGG network. 
Finally, to enforce temporal consistency we also add a sequence adversarial loss. This makes sure that the movement of the lips look realistic across frames. 
\begin{equation}
\label{eq:L_gan}
\begin{aligned}
\mathcal{L}_{\text{GAN}}=\mathbb{E}_{x_{0,s},t,\epsilon}[\log D_\psi (x^\theta_{0,s:s+5})]+ \\
\mathbb{E}_{x_{0,s}}[\log(1-D_\psi(x_{0,s:s+5}))]
\end{aligned}
\end{equation}
where we use a PatchGAN~\cite{patch-gan} discriminator $D_\psi$. This task requires more context than just two frames~\cite{chan2019everybody} but no optical flow~\cite{wang2018video}.

The overall optimization objective can be written as:

\begin{equation}
\label{eq:L_all}
\begin{aligned}
\mathcal{L}=\mathcal{L}_{\text{simple}} + \lambda_{\text{l2}}\mathcal{L}_{2} +  \lambda_{\text{sync}}\mathcal{L}_{\text{sync}}+ \\ \lambda_{\text{lpips}}\mathcal{L}_{\text{\text{lpips}}}+\lambda_{\text{gan}}\mathcal{L}_{\text{GAN}}
\end{aligned}
\end{equation}

For sequence-based losses, it is essential that the diffusion process step input $t$ is the same for a sequence of frames $x_{0,s:s+5}$. This ensures uniformity within a predicted sequence during loss computation.

%% file: 4_experiments.tex
\section{Experiments}\label{sec:exp}

\textbf{Datasets.} We evaluate our method on the Voxceleb2~\cite{voxceleb2} and LRW~\cite{lrw} datasets, which contain in-the-wild videos of talking human faces and are  commonly used for lip-sync research. 

\noindent{\textbf{Voxceleb2}}~\cite{voxceleb2} - consists of over 1M face-cropped Youtube videos coming from 6000+ identities. This dataset consists of high variation in lighting, image quality, pose, and motion blur. The average video length is ~8 seconds.

\noindent{\textbf{LRW}}~\cite{lrw} - is a lip-reading dataset that contains 1000 videos each of 500 different words for a length of 1 second coming from BBC news. It has less variation compared to Voxceleb2 and focuses on clean front-facing videos. 
Like previous works, our model is trained only on the Voxceleb2 train split while we test on both datasets. We don't use the whole dataset for training but rather only use the first utterance of every video, which totals 145K videos.

\textbf{Implementation Details.} We preprocess the videos to have a framerate of 25 fps and an audio sample rate of 16kHz. For all our models the video resolution is 224$\times$224 out of which we crop the face and resize it to 128$\times$128. This is then masked in the lower half using gaussian noise and fed to our model which only morphs the lower half of this image according to the audio input. Then we resize it back to the original crop size and place it back on the video. For audio inputs, we first sample the audio at 16kHz and then create mel-spectrograms with window-size 800 and hop-size 200. These audio frames turn out to have size 16$\times$80. We build our code on top of the guided-diffusion repository~\cite{guided-diffusion}. %
We train our model on 8 NVIDIA RTXA6000 GPUs which takes around 4 days. Our model is trained using $T=1000$ diffusion steps, but for faster inference, we use only 25 steps of DDIM~\cite{ddim} sampling which takes 4.67 seconds on an average for all the frames of one VoxCeleb2~\cite{voxceleb2} video (avg. 8 seconds at 25 fps) on 8 NVIDIA RTXA6000 GPUs.

\begin{table}[!t]
\setlength{\cmidrulewidth}{0.01em}
\renewcommand{\tabcolsep}{3pt}
\renewcommand{\arraystretch}{1.1}
\caption{Ablation over our losses (Reconstruction)}\label{tab:ablation_recon}
\centering
\resizebox{\linewidth}{!}{
\begin{tabular}{@{}lcccccc@{}}
\toprule
Losses & FID ↓ & SSIM ↑& PSNR ↑ & LMD ↓ & Sync$_\text{c}$ ↑  \\
 \midrule
Reconstruction          & 8.589          & 0.523          & 18.234         & 3.472          & 0.633         \\
+ SyncNet     & 8.998          & 0.526 & \textbf{18.57} & 3.123          & 6.336         \\
+ Perceptual & \textbf{7.751} & 0.526          & 18.548         & 3.121          & 6.53          \\
+ Seq. GAN           & 8.213 & \textbf{0.527} & 18.52 & \textbf{3.101} & \textbf{7.89} \\
\bottomrule
\end{tabular}
}
    \vspace{-0.2in}
\end{table}

\begin{table*}[!t]
\centering
\small
\setlength{\cmidrulewidth}{0.01em}
\renewcommand{\tabcolsep}{4pt}
\renewcommand{\arraystretch}{1.1}
\caption{Quantitative comparison with baselines on Voxceleb2~\cite{voxceleb2} and LRW~\cite{lrw} on the task of reconstruction and Cross generation.}
\label{tab:main_table}
\begin{tabular}{@{}llcccccccccc@{}}
\toprule
\multirow{2}{*}{Dataset} &
\multirow{2}{*}{Method} & \multicolumn{6}{c}{Reconstruction} & \multicolumn{4}{c}{Cross} \\
\cmidrule[\cmidrulewidth](l){3-8} \cmidrule[\cmidrulewidth](l){9-12}
& & FID ↓ & SSIM ↑& PSNR ↑ & LMD ↓ & Sync$_\text{c}$ ↑ & Sync$_\text{d}$ ↓ & FID ↓ & LMD ↓ & Sync$_\text{c}$ ↑ & Sync$_\text{d}$ ↓ \\
 \midrule

\multirow{3}{*}{VoxCeleb2} & Wav2Lip~\cite{wav2lip} & 3.26 & 0.53   & 18.18                        &  3.16 & \textbf{9.08} & 5.93               & 5.11               & 4.84               & \textbf{8.12} & 6.74              \\ 
 & PC-AVS~\cite{pcavs} & 4.25                        & 0.53               & \textbf{18.26}               & 3.16                        &  6.71  & 7.80 &  10.62 &  5.00 &  6.96 & 7.53\\ 
& \ours~(Ours) & \textbf{2.46}               & 0.53 &  18.09 & \textbf{3.04}               & 8.78           & \textbf{5.93}             & \textbf{4.53}                & \textbf{4.82}               &  7.62  & \textbf{6.73}  \\ 

\midrule
\multirow{3}{*}{LRW} & 
Wav2Lip~\cite{wav2lip}  &  4.23 & \textbf{0.68} & \textbf{20.76} & \textbf{2.15} & \textbf{8.13}  & \textbf{6.09 }         & 5.19 & \textbf{3.88} & \textbf{7.52} & \textbf{6.56}\\ 
 & PC-AVS~\cite{pcavs}     & 6.80 & 0.61 & 20.10 & 2.29 & 6.68 & 7.29
 & 8.48 & 4.09 & 6.66 & 7.27 \\ 
 & \ours~(Ours)            & \textbf{2.62} & 0.67 & 20.62 & 2.17 & 7.41 & 6.21
 & \textbf{2.54} & 3.93 & 6.44 & 6.97\\ 
\bottomrule
\end{tabular}

    \vspace{-0.1in}
\end{table*}

\textbf{Comparison Methods.} We compare our method against the most popular methods for lip-sync. Our choice of models is based also on models/codebases which are publicly available. Wav2Lip~\cite{wav2lip} is an inpainting style method that uses SyncNet expert loss to get good lip-sync. PC-AVS~\cite{pcavs} is a head reconstruction method that focuses on controlling pose apart from identity and lip shape. For both these methods we use their publicly available pre-trained models for the evaluation of all the datasets.

\begin{table}[!t]
\begin{minipage}[t]{\linewidth}
\setlength{\cmidrulewidth}{0.01em}
\renewcommand{\tabcolsep}{9pt}
\renewcommand{\arraystretch}{1.1}
\caption{Ablation over our losses (Cross generation)}\label{tab:ablation}
\centering
\resizebox{0.8\linewidth}{!}{
\begin{tabular}{@{}lcccc@{}}
\toprule
Losses & FID ↓ & LMD ↓ & Sync$_\text{c}$ ↑ \\
 \midrule

Reconstruction          & 6.694          & 5.313          & 0.992         \\
+ SyncNet     & 8.784          & \textbf{4.816} & 5.946         \\
+ Perceptual & 5.016          & 5.009          & 5.955          \\
+ Seq. GAN           & \textbf{4.592} & 4.985          & \textbf{6.83} \\
\bottomrule
\end{tabular}
}
\end{minipage}

    \vspace{-0.1in}
\end{table}

\begin{figure*}[!t]
    \centering
    \label{fig:qual_recon}
    \includegraphics[width=0.9\linewidth]{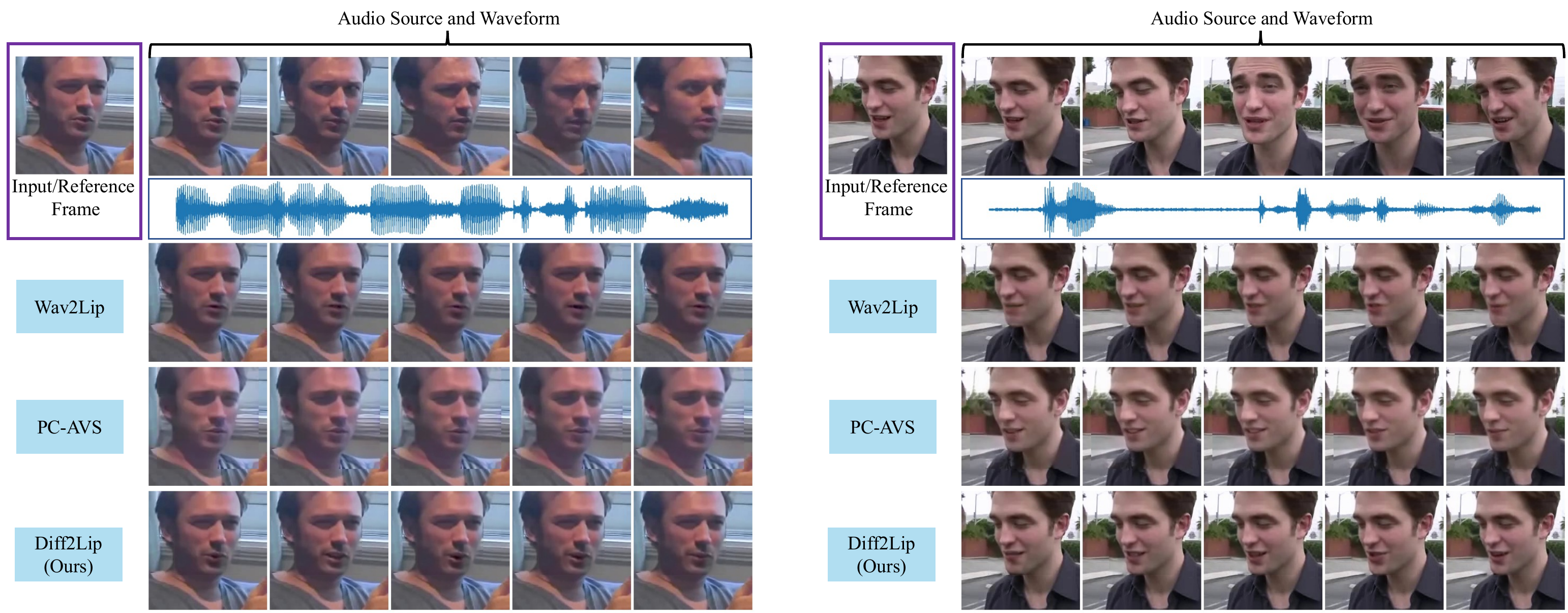}
    \caption{\textbf{Qualitative results of Reconstruction on VoxCeleb2~\cite{voxceleb2}}. Here we provide only the first frame as the input source (for pose) as well as the reference frame (for identity), and this frame is driven using the audio (second row) coming from the same video (top row). Wav2Lip~\cite{wav2lip} blurs the lip region in both cases to achieve the correct lip shape while PC-AVS has identity loss (see right) and border discontinuity. Our generations look highly realistic and have the lip shapes as in the audio source. (Please zoom in for better visibility.)}
    \label{fig:qualitative-recon}
    \vspace{-0.1in}
\end{figure*}

\begin{figure*}[!h]
    \centering
    \label{fig:qual_cross}
    \includegraphics[width=0.9\linewidth]{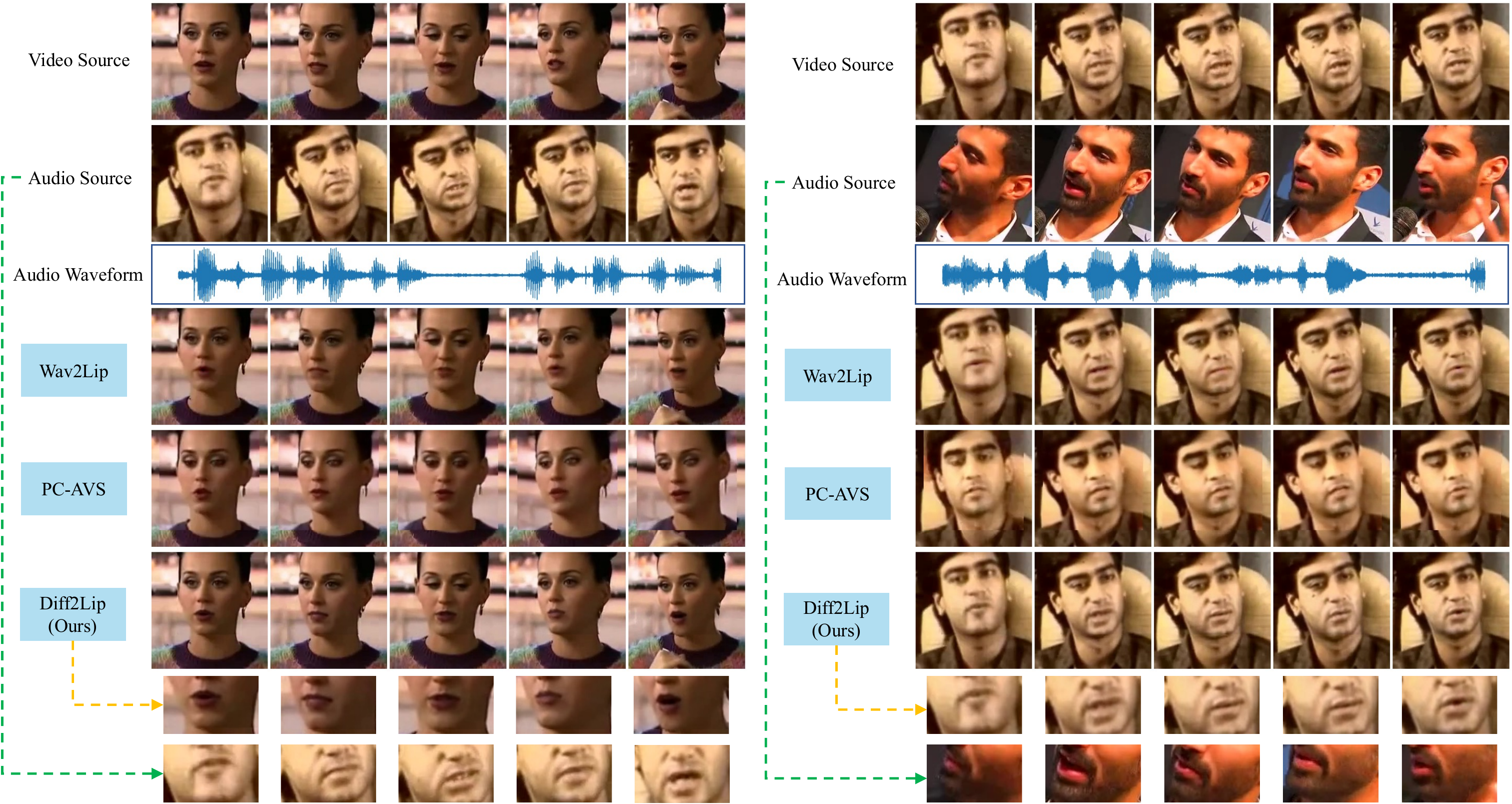}
    \caption{\textbf{Qualitative results of Cross generation on VoxCeleb2~\cite{voxceleb2}}. Here we provide a video source (first row) and drive that identity-pose combination using audio coming from a different video (second and third rows). Wav2Lip~\cite{wav2lip} blurs its generations, for example, beard region details are missing on the right. PC-AVS's~\cite{pcavs} generations have flaws like identity loss, in both cases. They introduce artifacts near the eyes on the left while there is identity loss on the right. Our method generates realistic mouths with expressive lips while being in sync with the audio source. In the bottom 2 rows, we can see that the lip region of our generations match those of the audio source. (Please zoom in for better visibility.)}
    \label{fig:qualitative-cross}
    \vspace{-0.1in}
\end{figure*}

\subsection{Quantitative Evaluation}\label{sec:quant-eval}
 For quantitative evaluation, we evaluate our model in terms of both the visual quality as well as audio synchronization. For visual quality, we use FID~\cite{fid}, SSIM~\cite{ssim}, and PSNR, which are popular metrics used in papers like Wav2Lip~\cite{wav2lip}, PC-AVS~\cite{pcavs}, AV-CAT~\cite{avcat}. FID is a popular metric used for comparing the ``realness'' of generated images by comparing against the real image distribution. SSIM and PSNR are pixel-wise image similarity metrics that compare a pair of images and are not suited for capturing variability in video generation~\cite{shrivastava2021diverse} but are included in this work for completeness. While to measure synchronization we use LMD~\cite{chen2018lip}, $\text{Sync}_\text{c}$, and $\text{Sync}_\text{d}$~\cite{syncnet}. LMD measures the distance between mouth landmarks among frames. $\text{Sync}_{c}$ is the confidence score of SyncNet while $\text{Sync}_\text{d}$ is the average distance between SyncNet video and audio representations, which tell the synchronization quality. Note that for evaluation, we use the pre-trained SyncNet from the SyncNet's~\cite{syncnet} repository but for training, we train our own SyncNet, similar to Wav2Lip\cite{wav2lip} and AV-CAT\cite{avcat}.
We use pre-trained models of Wav2Lip and PC-AVS methods to conduct our evaluations. Wav2Lip has provided its code for calculating FID and $\text{Sync}_\text{c}$ and we use the same for these metrics. We use face-alignment ~\cite{face-alignment} for landmark detection and the LMD metric proposed in~\cite{chen2018lip}. For SSIM and PSNR we use the same inputs as used for FID, consequently, our values are a bit different compared to~\cite{pcavs}. This is  possibly because they evaluate these metrics at different scales and lack these evaluation details. On the other hand, we tend to get better LMD and $\text{Sync}_\text{c}$ values for PC-AVS than noted in their paper. Some papers like~\cite{gcavt} show PC-AVS's metrics only for reference as its generations occasionally fail in landmark detection. For uniformity of scale, we uncropped PC-AVS's generations and paste them back on the background before evaluation.

\begin{figure*}[!t]
    \centering
    \includegraphics[width=0.95\linewidth]{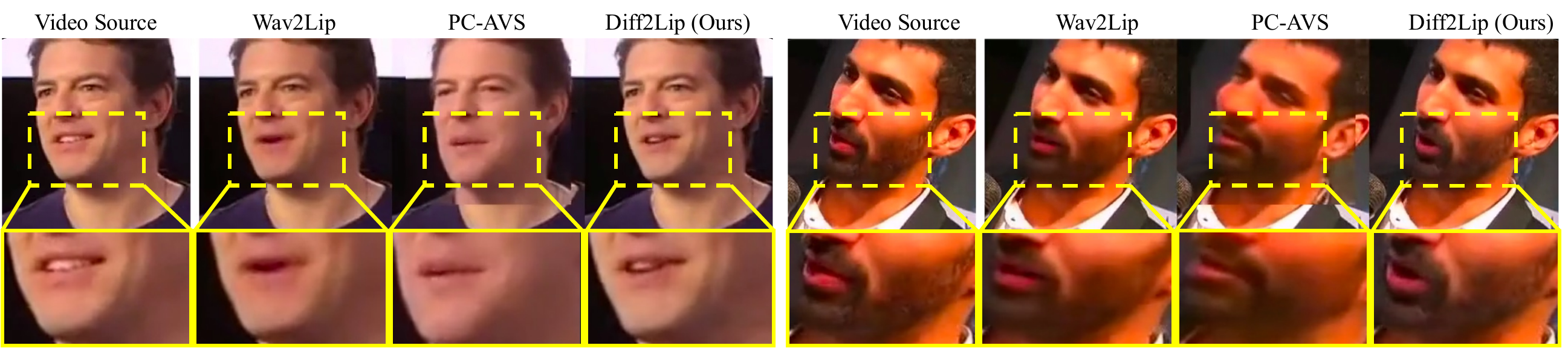}
    \caption{\textbf{Qualitative Visual-quality Comparison}. We zoom in on the mouth region of two examples and compare them against the video source. Wav2Lip~\cite{wav2lip} blurs the lip region while PC-AVS~\cite{pcavs} tends to change the identity. \ours~ preserves identity and generates high-fidelity lips.}
    \label{fig:qualitative-zoomed}
    \vspace{-0.1in}
\end{figure*}

\subsubsection{Reconstruction}\label{sec:recon}
Similar to the setting mentioned in~\cite{pcavs} and later used in~\cite{gcavt} and~\cite{avcat}, we evaluate the methods on the task of reconstruction of the video given only the first frame and the audio corresponding to the same video. The audio is used as the driver for this reconstruction. Note that PCAVS requires an additional pose input apart from the identity input. We feed the first frame to it for both inputs, similar to the ``fixed pose'' setting in their paper. For Wav2lip and ours, the first frame acts as both the input frame and the reference frame. For Voxceleb2, we show the results on 4911 test utterances instead of 35K videos due to resource constraints. These 4911 utterances are the first utterance of each video in the test set, and hence cover all the videos in the test set. For LRW, our results are noted on all 25K videos in the test set.

The results are noted in Table~\ref{tab:main_table}. We observe that \ours~outperforms both the methods with respect to FID metric on both datasets which points towards better generation quality. The SSIM, PSNR and LMD values of our method are comparable with the other methods. We see that Wav2Lip's  $\text{Sync}_\text{c}$ is better than both ours and PC-AVS's. This is possibly because our SyncNet expert may be weaker in performance compared to the one used in Wav2Lip. 

\vspace{-0.1in}
\subsubsection{Cross generation}\label{sec:cross}
\vspace{-0.1in}
We also evaluate the method on the task of lip-sync when the identity and the pose are controlled using a video while the lip-sync is driven using input audio corresponding to a different video. This was introduced in~\cite{wav2lip} and is a more realistic setting as here the generations are closer to lip-sync in-the-wild. We use the input frame as the reference frame in this setting similar to Wav2Lip, as that provides the best texture information of the frame. For PC-AVS, the input frames are fed as both pose and identity sources. Note that we cannot evaluate SSIM and PSNR for this setting because there are no ground truth frames available. So, we provide the rest of the metrics in Table~\ref{tab:main_table}. For Voxceleb2, we select 4970 pairs of audio-video combinations where the two sources are different. We sample these using all the pair combinations of the first utterance of the first video coming from 71 randomly chosen test identities. For LRW, we use the 28K audio-video pair provided in Wav2Lip's~\cite{wav2lip} evaluation.  
 The results are noted in Table~\ref{tab:main_table}. Similar to reconstruction evaluation, we here as well observe that our method excels in image quality while being comparable in other metrics except for $\text{Sync}_\text{c}$.

\subsection{Qualitative Evaluation}
For qualitative evaluation, we show visually compare against on both reconstructions (in Fig.~\ref{fig:qualitative-recon}) as well as cross generation (in Fig.~\ref{fig:qualitative-cross}). These settings are the same as introduced in Section~\ref{sec:recon} and \ref{sec:cross}. It can be observed in these qualitative results that PC-AVS tends to lose the identity of the source video and also suffers from boundary discontinuity problems which make it unsuitable for in-the-wild generation. On the other hand, Wav2Lip tends to generate blurred-out mouth regions so as to achieve good lip sync. \ours~ does not suffer from these issues and is able to generate high-quality mouth region while having expressive lip shapes which correctly correspond to the ground truth (audio sources' mouth shape) as seen in Fig.~\ref{fig:qualitative-zoomed}.

\begin{table}[!t]
\small
\setlength{\cmidrulewidth}{0.01em}
\renewcommand{\tabcolsep}{3pt}
\renewcommand{\arraystretch}{1.1}
\caption{User Study measured by Mean Opinion Scores~(MOS) (max. 5) and  Preference in percentage.}\label{tab:user_study}
\centering
\begin{tabular}{@{}lccc@{}}
\toprule
Measure & Wav2Lip & PC-AVS  & \ours \\
 \midrule

MOS (Visual quality) ↑   & 3.75 & 2.71 & \textbf{4.16} \\
MOS (Lip-sync quality) ↑ & 3.84 & 3.34 & \textbf{3.86} \\
MOS (Overall quality) ↑  & 3.70 & 2.91 & \textbf{3.91} \\
Preference ↑         & 37\% & 8.33\% & \textbf{54.67}\% \\
\bottomrule
\end{tabular}
    \vspace{-0.1in}
\end{table}

\textbf{User Study.} We conduct a user study where we ask 15 participants to judge lip-sync videos generated in cross generation setting by \ours~ and two other methods. 20 videos were sampled from the VoxCeleb2's test set and are driven by randomly selected driving audios. The participants rated the videos 1-5 (where higher is better) in the aspects of (1) \textbf{Visual quality} (2) \textbf{Lip-sync quality}, and (3) \textbf{Overall quality}. We used the Mean Opinion Score (MOS) measure to aggregate these ratings. Further, we record the percentage of times users preferred a method. We present the results in Table~\ref{tab:user_study}, where we see that our method surpasses others in all the categories. In terms of Lip-sync quality, this is opposed to our quantitative results, especially $\text{Sync}_\text{c}$. We speculate that $\text{Sync}_\text{c}$, might favor blurry generations with high temporal consistency while humans prefer high fidelity over slight temporal inconsistency.

\subsection{Ablations}
We conduct an ablation study to showcase the contribution of various choices of losses that were used during training. Specifically, we train our model in three different settings in which we introduce an additional loss in each setting. First, in the setting, we train our model using only $\mathcal{L}_\text{simple}$ (Reconstruction). Second, we train another version of the model 
 using $\mathcal{L}_\text{simple}+\mathcal{L}_{2}+\mathcal{L}_\text{sync}$ (+ SyncNet). Here intuitively the $\mathcal{L}_\text{sync}$ should introduce better synchronization. Third, we further add a perceptual loss $\mathcal{L}_\text{sync}$ (+ Pecept). We add this loss because adding the SyncNet loss led to worse image quality. Finally, we add the sequential adversarial loss $\mathcal{L}_\text{GAN}$ to achieve even temporal consistency(+ Seq. GAN). We test these on a smaller subset of 500 VoxCeleb2 test audio-video pairs in the cross generation setting as well as the reconstruction setting. 
 
It can be seen in Table~\ref{tab:ablation} that moving from ``Reconstruction" to ``+ SyncNet" gives a sudden improvement in the $\text{Sync}_\text{c}$ metric. This supports our intuition that only reconstruction-based losses are not enough. We also see that this transition deteriorates the image quality. This gets solved as we move to the ``+ Perceptual" setting. Finally, the addition of sequential adversarial loss not just further improves the image quality but also improves the $\text{Sync}_c$, clearly showing the advantage of this loss.  In the reconstruction setting in Table~\ref{tab:ablation_recon}, most of these observations still hold except FID being lower for ``+ Perceptual" than ``+ Seq. GAN". This could be attributed to the static nature of the input source in this setting while the ground truth is moving.

%% file: 5_discussion.tex
\section{Discussion and Conclusion}
 \vspace{-0.1in}
\noindent\textbf{Discussion.} 
Even though \ours, cannot be used for facial reenactment and hence cannot be used for harmful acts like face-swapping, it can be used for other malicious purposed like disinformation. We discourage and disapprove of any such applications which may have negative implications and strictly condone their use for positive purposes.

\noindent\textbf{Conclusion.} In this work, we present \ours, which is able to generate high-quality lip synchronization. We pose the task to be a mouth region inpainting task and solve it by learning an audio-conditioned diffusion model. Our ablation studies show that SyncNet loss is required in our framework to introduce lip-sync while sequential adversarial loss improves both image quality and temporal consistency. Finally, extensive quantitative and qualitative results validate that our method performs better than state-of-the-art methods in terms of image quality while maintaining other metrics and also being preferred by the users.

\noindent\textbf{Acknowledgements.} 
We would like to thank our colleagues 
Matthew Gwilliam, Archana Swaminathan, and Ahmed Taha for their feedback on this work and all the participants of the user study for their time.

%% file: 6_supp.tex
\iccvfinalcopy 

\ificcvfinal\pagestyle{empty}\fi

\title{
\Large
\noindent\textbf{Appendix}
}

\section{Denoising Diffusion Implicit Models (DDIM)}
In DDIM formulation~\cite{ddim}, the posterior is defined such that one can control its variance schedule $\{\sigma_t\in \mathbb{R}_{\geq0}\}^T_{t=1}$. Hence, in the reverse diffusion process, $x_{t-1}$ can be sampled from $x_t$ using:
 \begin{equation}
\label{eq:ddim}
x_{t-1}=\sqrt{\Bar{\alpha}_{t-1}}x_0 + \sqrt{1-\Bar{\alpha}_{t-1} - \sigma_t^2}\left(\frac{x_t-\sqrt{\Bar{\alpha}_{t}}x_0}{\sqrt{1-\Bar{\alpha}_{t}}} \right) \\
+ \sigma_t\xi
\end{equation}
for $\xi\in \mathcal{N}(0, \textbf{I})$. If for all $t$, $\sigma_t=\sqrt{(1-\Bar{\alpha}_{t-1})/(1-\Bar{\alpha}_{t})}\sqrt{1-\Bar{\alpha}_{t}/\Bar{\alpha}_{t-1}}$, this formulation becomes the same as standard DDPM~\cite{ddpm}. On the other hand, if for all $t$, $\sigma_t=0$, then the reverse diffusion process sampling becomes deterministic and simplifies to Eq. (4) from the main text. Further, in this formulation, one can use strided timesteps to make inference faster without training the model again. For example, instead of $t=1,2,3,...,1000$, if one uses $t=40, 80, 120, ... , 1000$, the inference can be done in 25 steps instead of 1000 steps.  

\section{Qualitative Results}
\subsection{More Qualitative Figures}
\begin{figure*}[!t]
    \centering
    \includegraphics[width=0.95\linewidth]{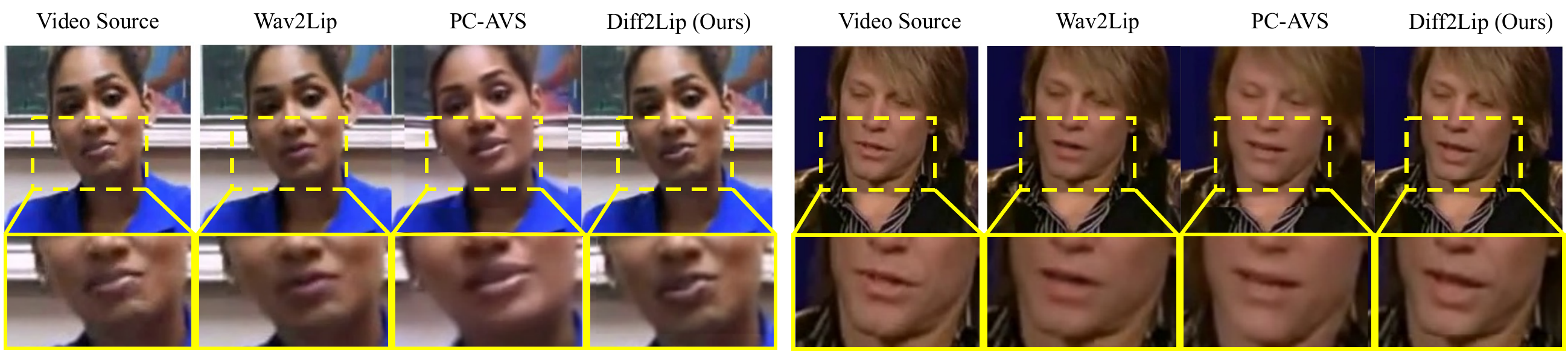}
    \includegraphics[width=0.95\linewidth]{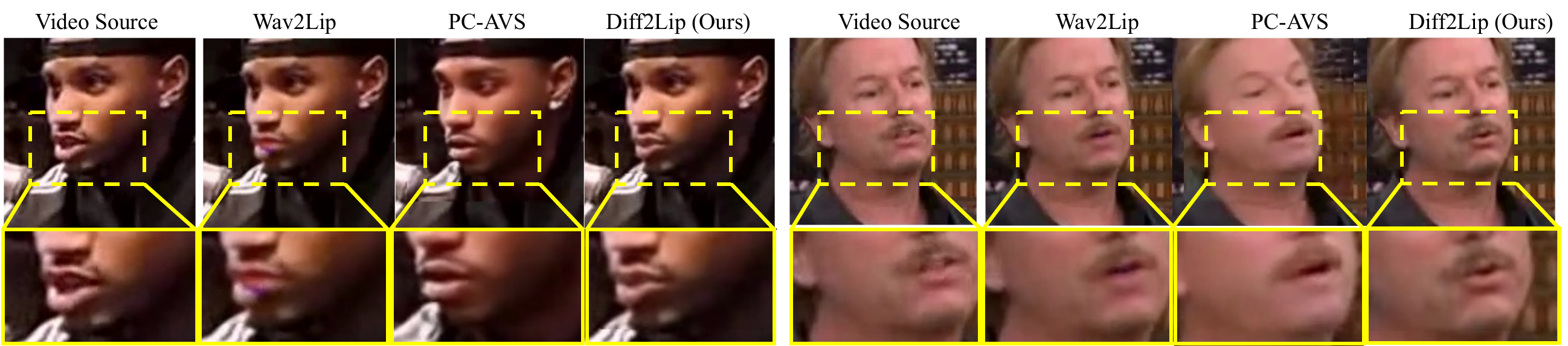}
    \caption{\textbf{Qualitative Visual-quality Comparison}. }
    \label{fig:qualitative-zoomed1}
    \vspace{-0.1in}
\end{figure*}

\begin{figure*}[!t]
    \centering
    \label{fig:qual_recon}
    \includegraphics[width=0.8\linewidth]{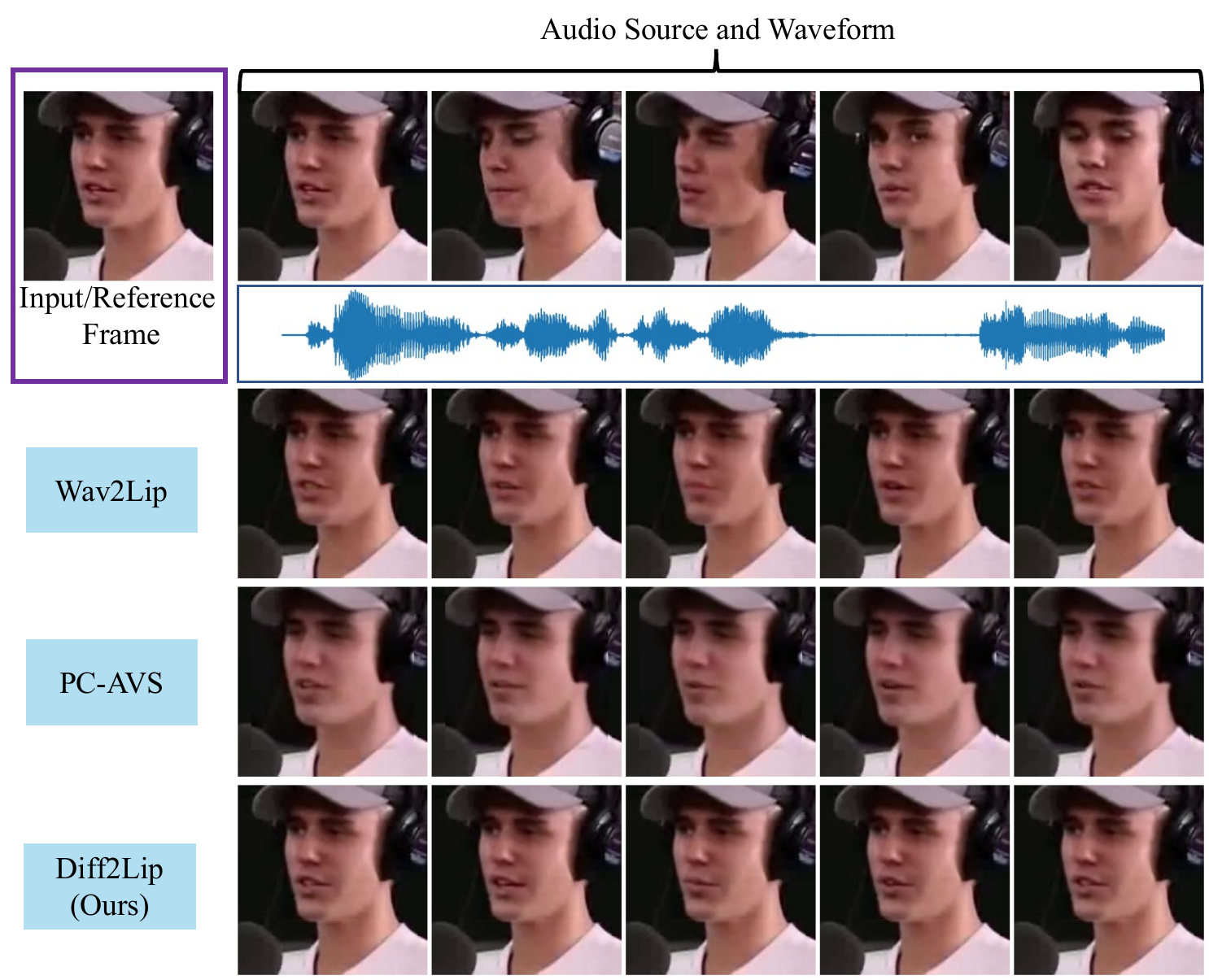}
    \includegraphics[width=0.8\linewidth]{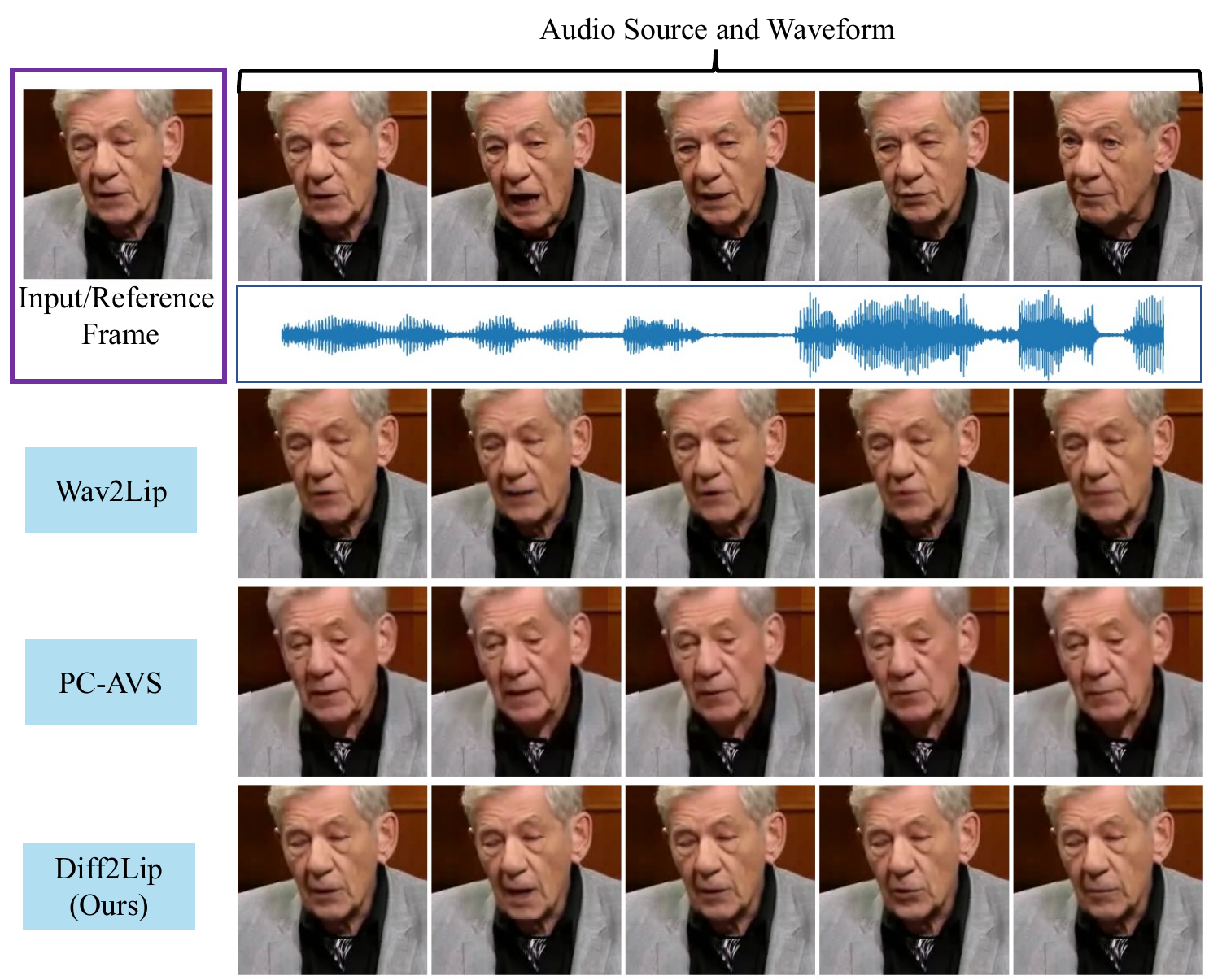}
    \caption{\textbf{Qualitative results of Reconstruction on VoxCeleb2~\cite{voxceleb2}}.}
    \label{fig:qualitative-recon1}
\end{figure*}

\begin{figure*}[!t]
    \centering
    \label{fig:qual_recon}
    \includegraphics[width=0.8\linewidth]{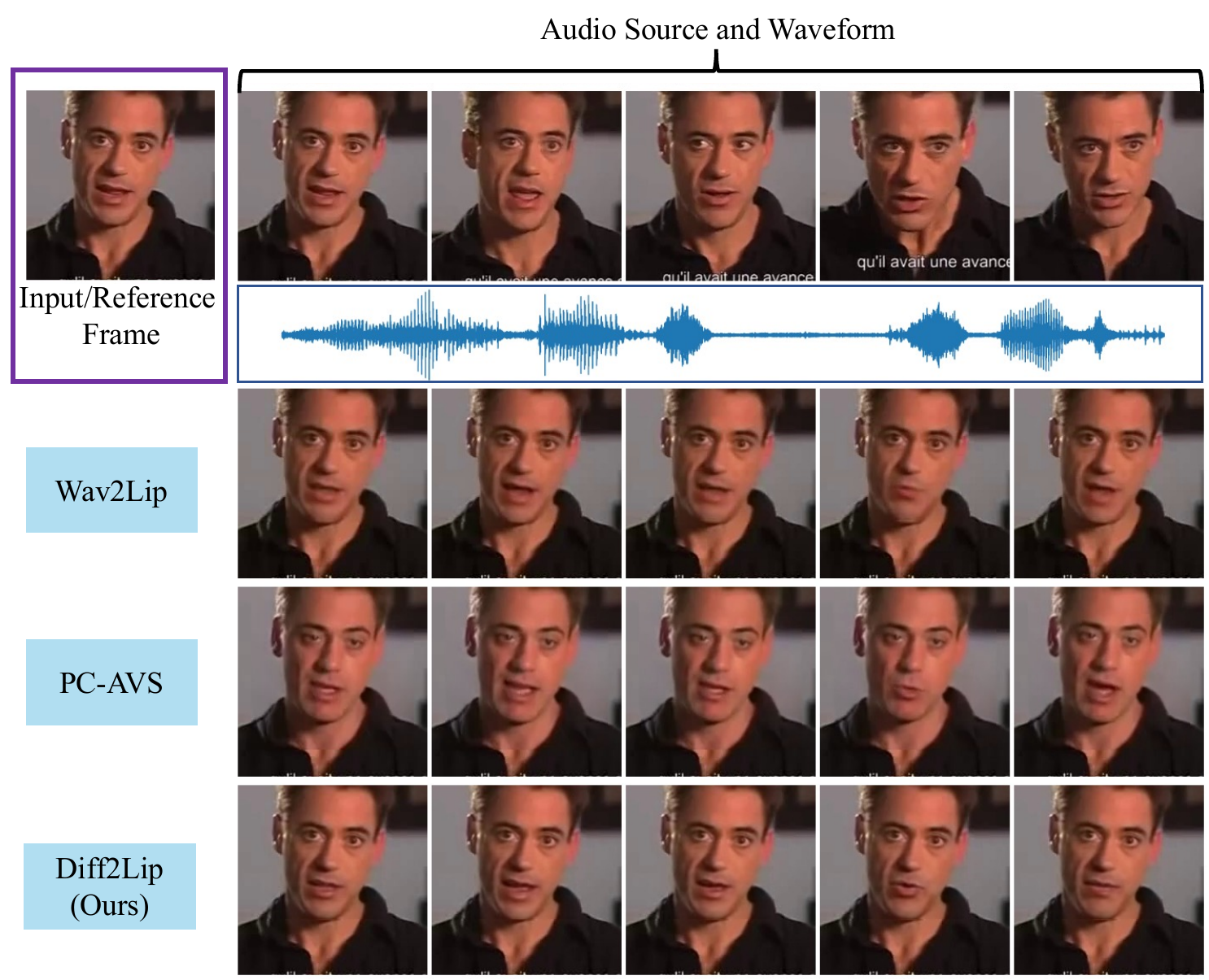}
    \includegraphics[width=0.8\linewidth]{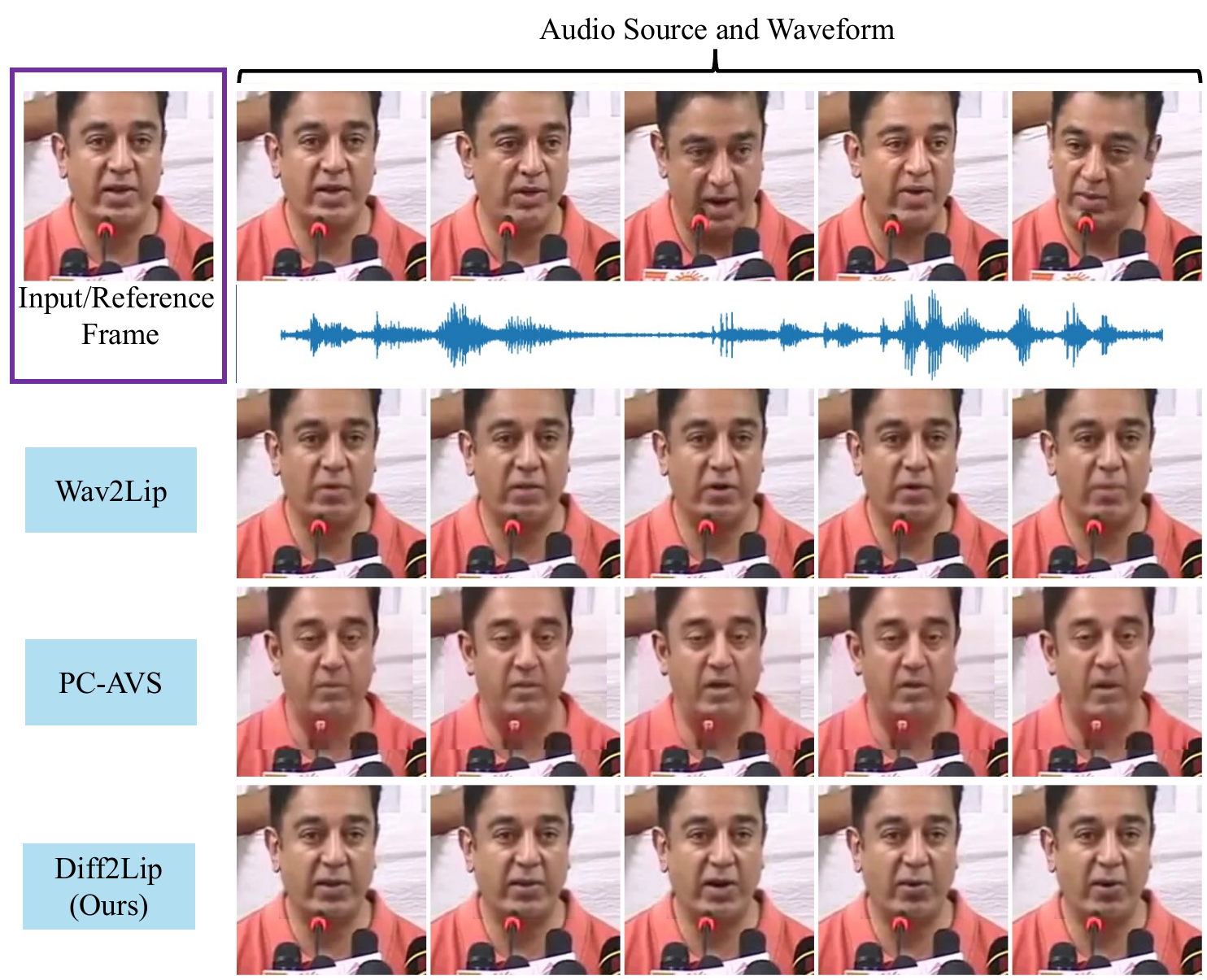}
    \caption{\textbf{Qualitative results of Reconstruction on VoxCeleb2~\cite{voxceleb2}}.}
    \label{fig:qualitative-recon2}
\end{figure*}

\begin{figure*}[!t]
    \centering
    \label{fig:qual_recon}
    \includegraphics[width=0.8\linewidth]{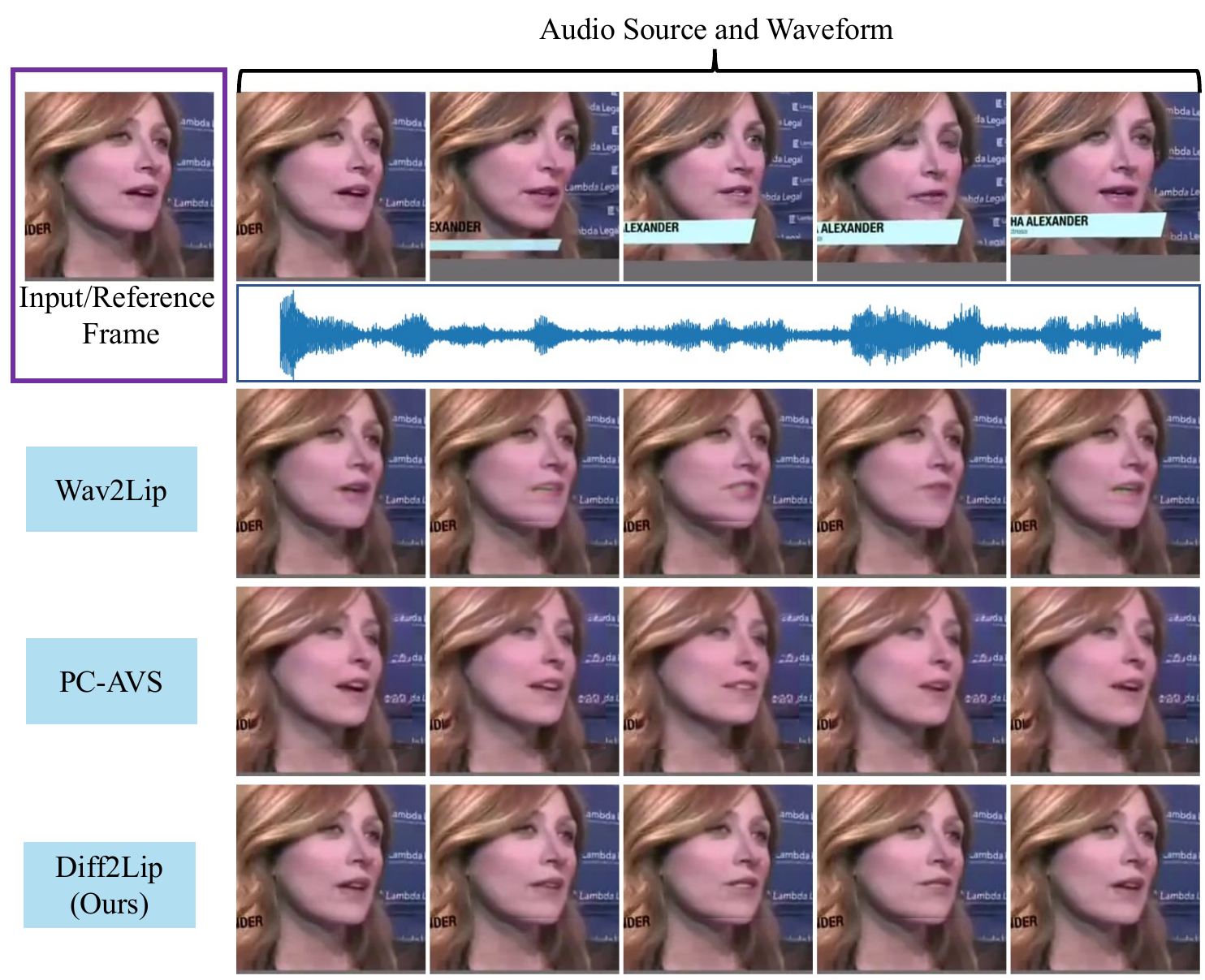}
    \includegraphics[width=0.8\linewidth]{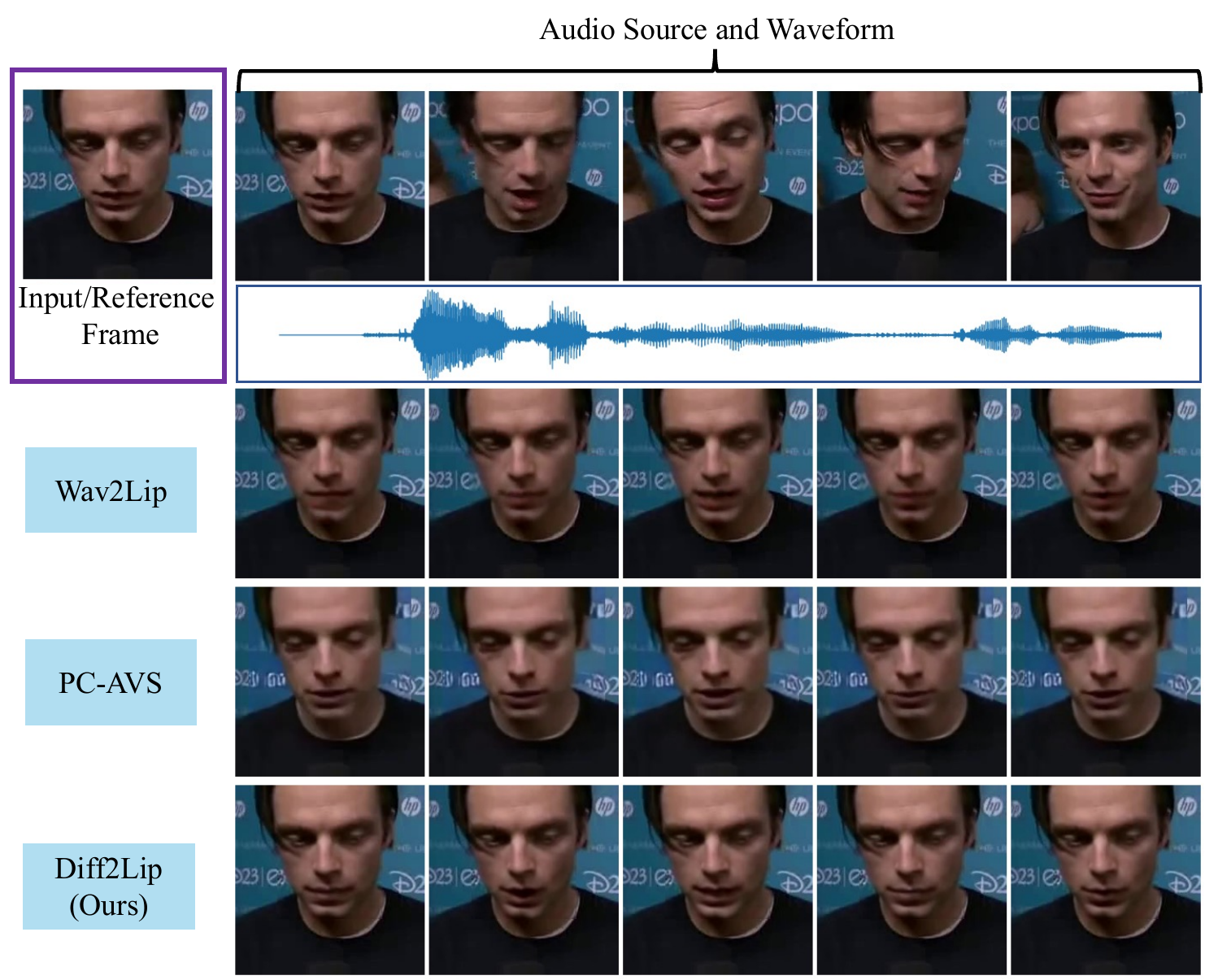}
    \caption{\textbf{Qualitative results of Reconstruction on VoxCeleb2~\cite{voxceleb2}}.}
    \label{fig:qualitative-recon3}
\end{figure*}

\begin{figure*}[!t]
    \centering
    \label{fig:qual_recon}
    \includegraphics[width=0.8\linewidth]{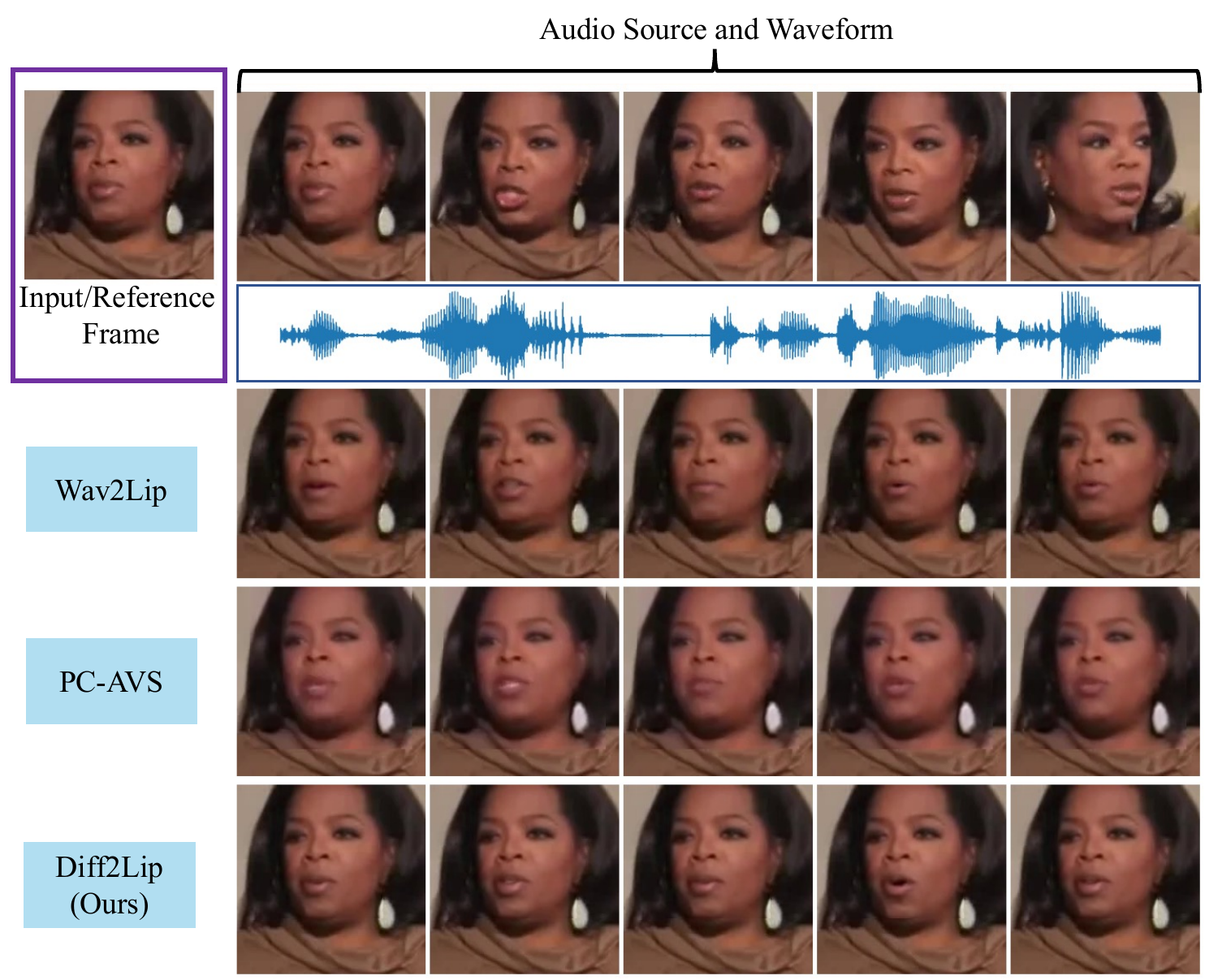}
    \caption{\textbf{Qualitative results of Reconstruction on VoxCeleb2~\cite{voxceleb2}}.}
    \label{fig:qualitative-recon4}
\end{figure*}

\begin{figure*}[!h]
    \centering
    \label{fig:qual_cross}
    \includegraphics[width=0.6\linewidth]{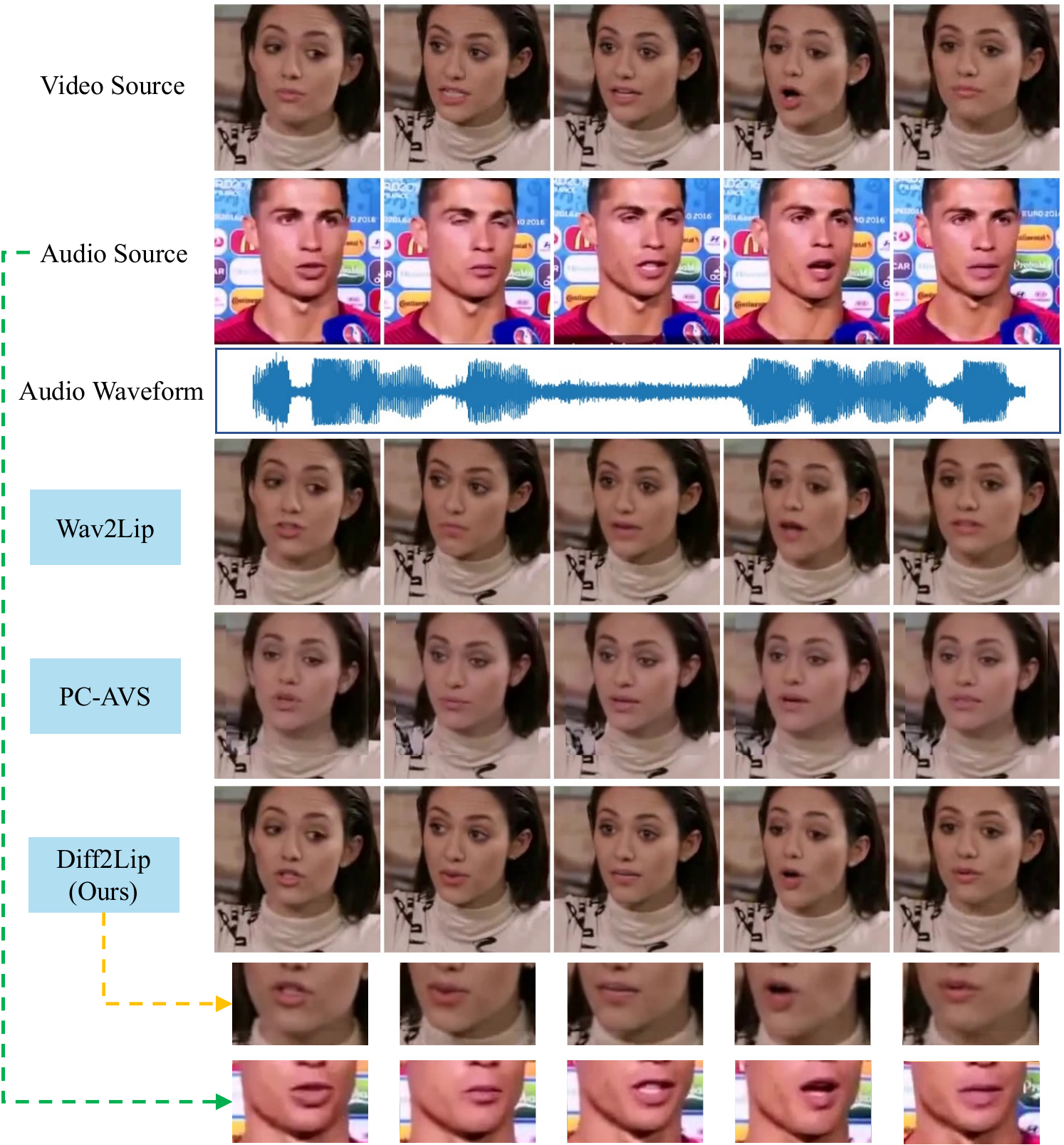}
    \\\noindent\makebox[\linewidth]{\rule{0.5\paperwidth}{0.4pt}}\\
    \includegraphics[width=0.6\linewidth]{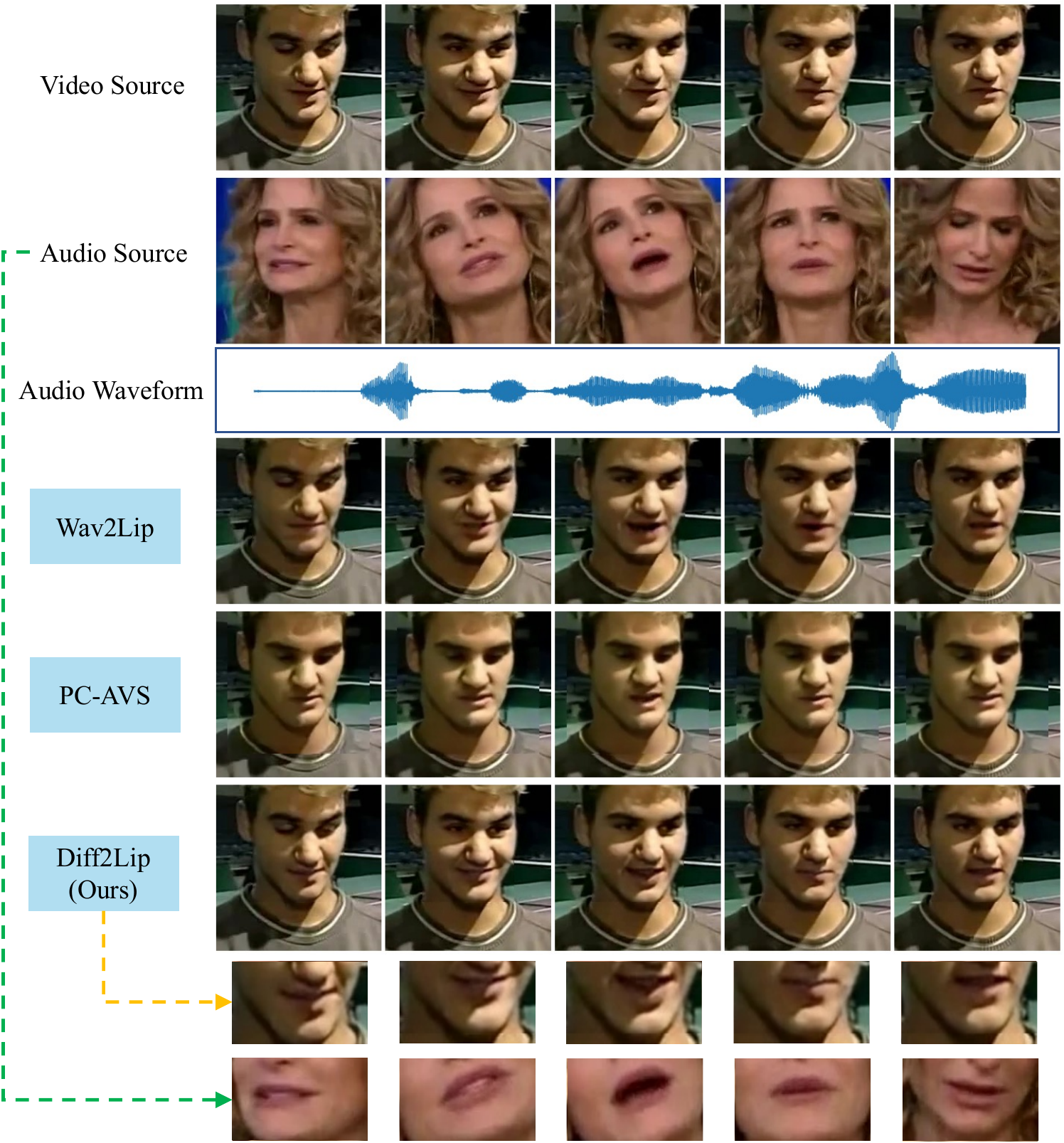}
    \caption{\textbf{Qualitative results of Cross generation on VoxCeleb2~\cite{voxceleb2}}.}
    \label{fig:qualitative-cross1}
    \vspace{-0.1in}
\end{figure*}

\begin{figure*}[!h]
    \centering
    \label{fig:qual_cross}
    \includegraphics[width=0.6\linewidth]{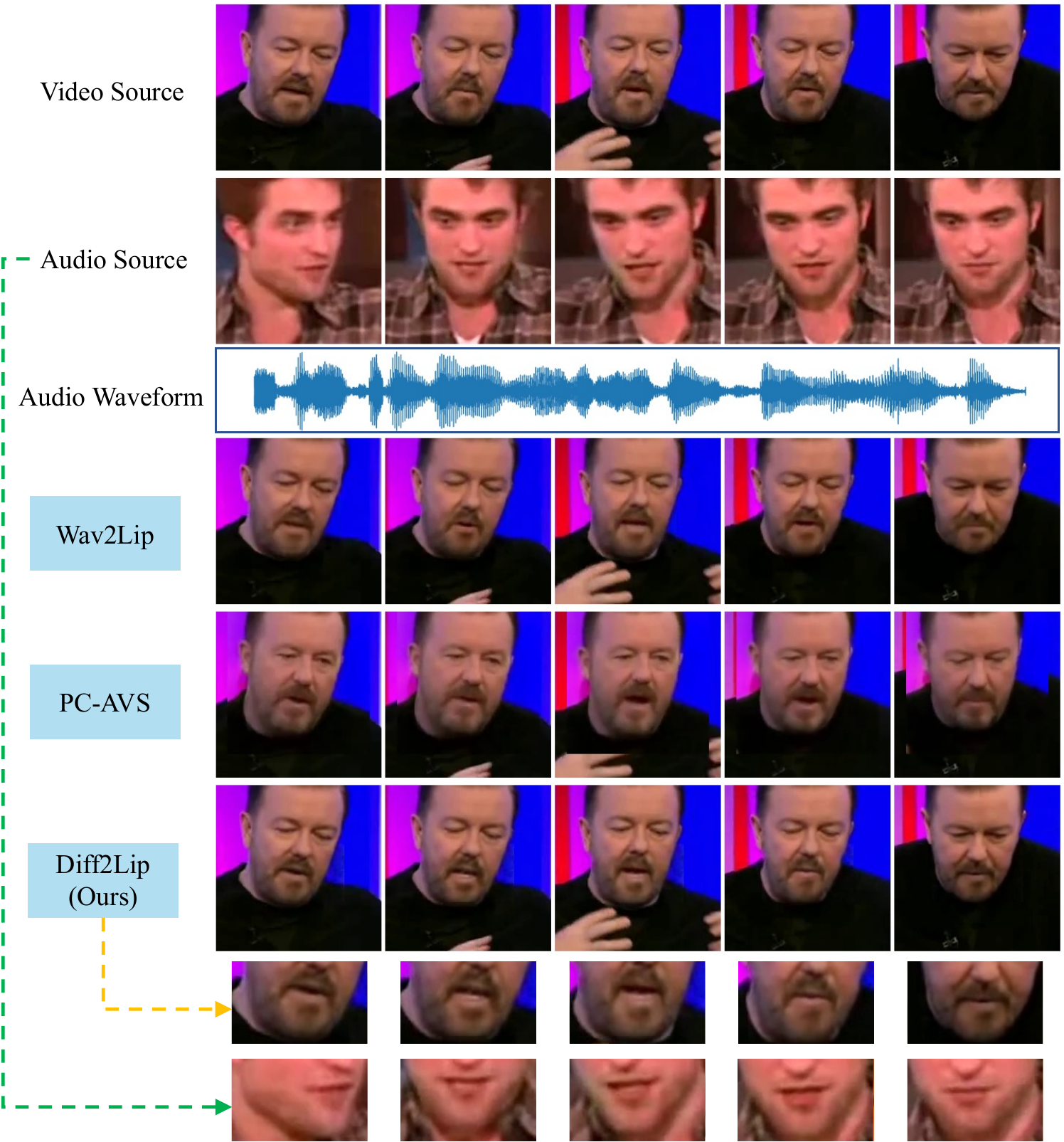}
    \\\noindent\makebox[\linewidth]{\rule{0.5\paperwidth}{0.4pt}}\\
    \includegraphics[width=0.6\linewidth]{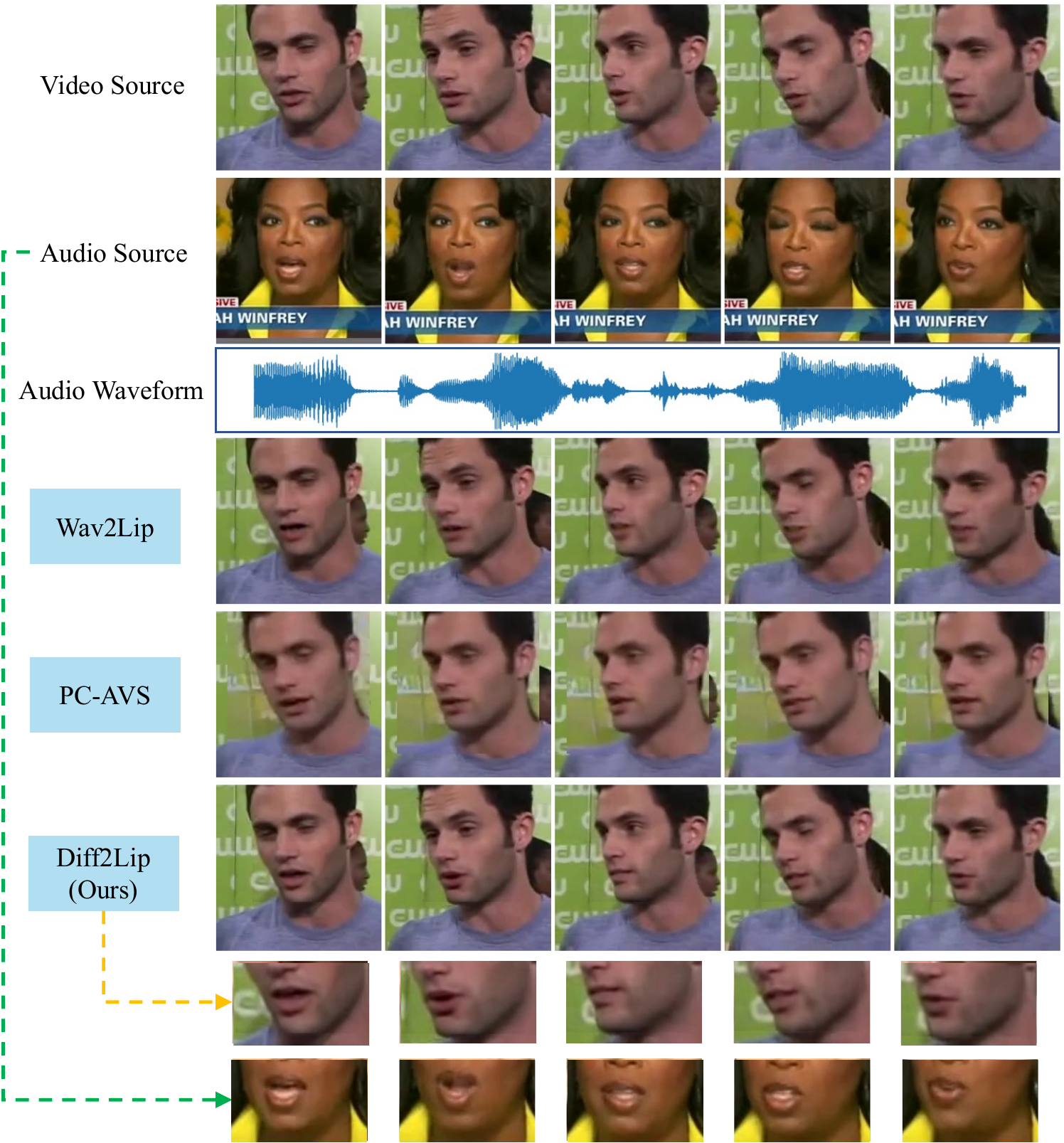}
    \caption{\textbf{Qualitative results of Cross generation on VoxCeleb2~\cite{voxceleb2}}.}
    \label{fig:qualitative-cross2}
    \vspace{-0.1in}
\end{figure*}

\begin{figure*}[!h]
    \centering
    \label{fig:qual_cross}
    \includegraphics[width=0.6\linewidth]{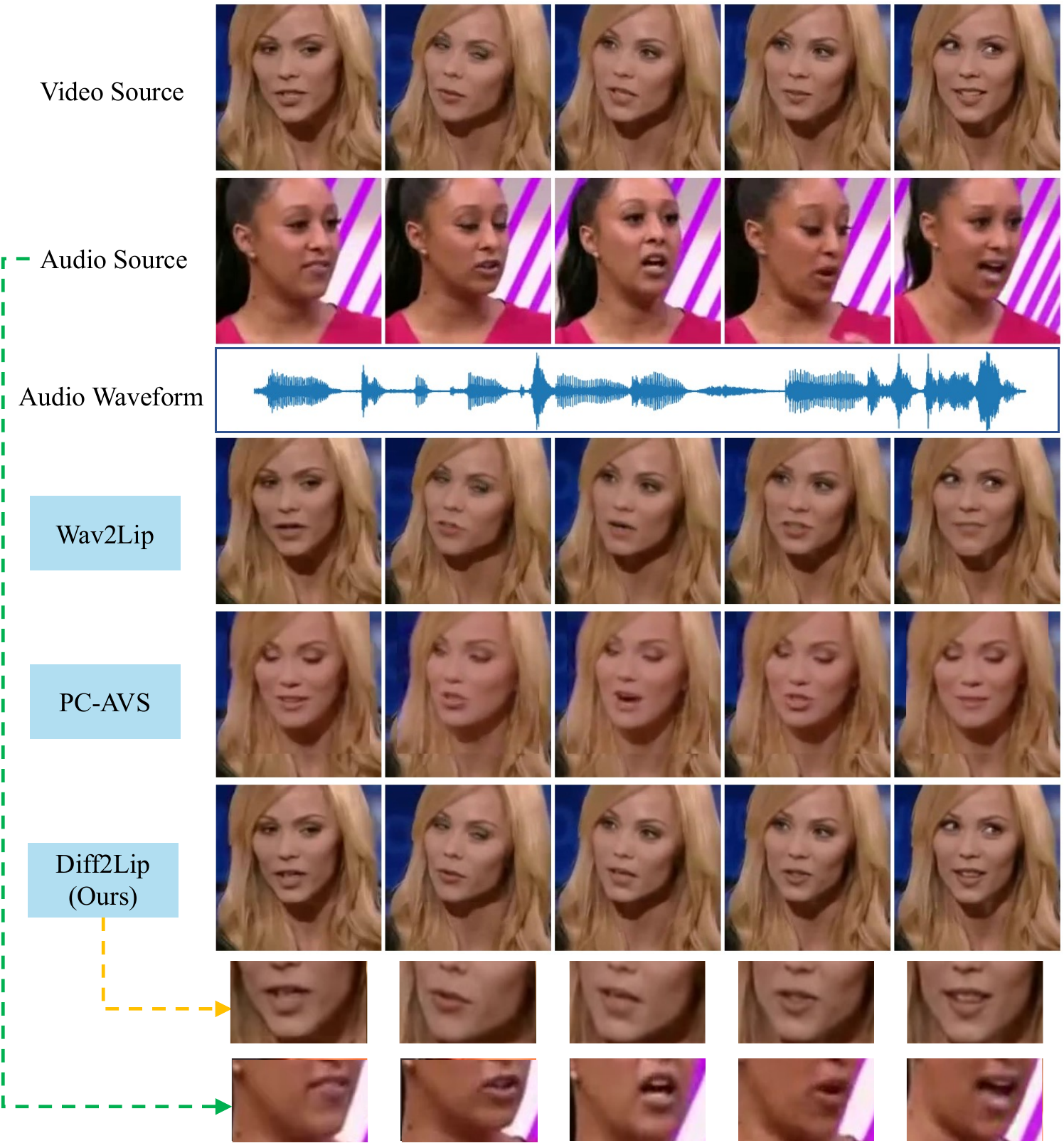}
    \\\noindent\makebox[\linewidth]{\rule{0.5\paperwidth}{0.4pt}}\\
    \includegraphics[width=0.6\linewidth]{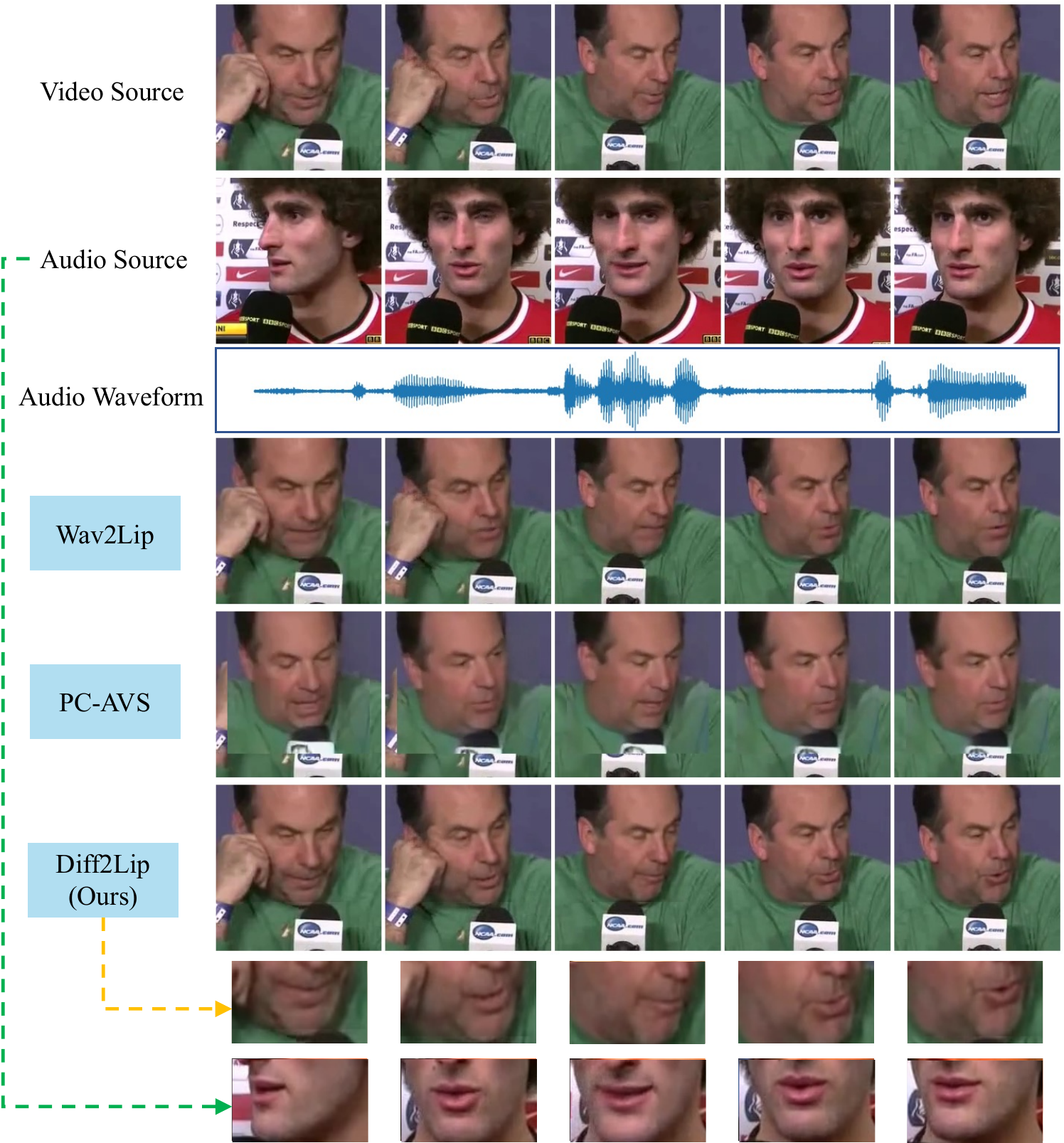}
    \caption{\textbf{Qualitative results of Cross generation on VoxCeleb2~\cite{voxceleb2}}.}
    \label{fig:qualitative-cross3}
    \vspace{-0.1in}
\end{figure*}

Similar to Fig.~\ref{fig:qualitative-zoomed} in the main paper, we show more visual comparisons in Fig. \ref{fig:qualitative-zoomed1}. Here we compare our approach with Wav2Lip~\cite{wav2lip} and PC-AVS~\cite{pcavs}. It can be seen in the zoomed-in lip regions that \ours~outperforms other methods both in image quality and identity preservation. Further, we show more examples of reconstruction setting in Fig. \ref{fig:qualitative-recon1}, \ref{fig:qualitative-recon2}, \ref{fig:qualitative-recon3}, and \ref{fig:qualitative-recon4} similar to Fig.~\ref{fig:qualitative-recon} in main paper. Here we can observe that our method produces the correct lip shape corresponding to the audio, which can be seen when compared with the original video (top row). Finally, Fig. \ref{fig:qualitative-cross1},  \ref{fig:qualitative-cross2},  \ref{fig:qualitative-cross3} show more results on cross generation setting similar to Fig.~\ref{fig:qualitative-cross} of the main paper. 

\subsection{Video Results}
Please find more video results at our \href{https://soumik-kanad.github.io/diff2lip}{project page}. 
We provide multiple pages of video result comparisons with other methods as well as the original video. The website also contains an interactive demo to see the intermediate stages of the diffusion process. We also show in the wild results using clips from movies dubbed in foreign languages. This along with the cross generations shows the generalizability of the proposed approach to unseen audios and identities. It also supports our motivation and vision to extend videos to different languages beyond simple dubbing which does not have synchronised lip movements. Our method is able to generate realistic lip-movements aligned with the words corresponding to new audio thus making the viewing experience better.

\section{Additional details}

We provide the filelists for the subsets of VoxCeleb2~\cite{voxceleb2} videos on which we have evaluated for both reconstruction and cross generation settings in our \href{https://github.com/soumik-kanad/diff2lip}{codebase}. Each line contains the relative path to the audio and the video separated by a space. For the case of reconstruction, both the audio and video paths are the same in a line. 


